# Optimized Quality of Service prediction in FSO Links over South Africa using Ensemble Learning.

S.O Adebusola [1], P. A Owolawi[1], J.S Ojo[2], P. S Maswikaneng[3]

[1]Department of Computer Systems Engineering, Tshwane University of Technology, Pretoria 0001, South Africa.

[2]Department of Physics, Federal University of Technology, Akure 340110, Nigeria.

[3]Department of Information Technology, Tshwane University of Technology, Pretoria 0001, South Africa

**ABSTRACT**: Fibre-optic communication system is expected to increase exponentially in terms of application due to the numerous advantages over copper wires. The optical network evolution presents several advantages such as over long-distance, low-power requirement, higher carrying capacity and high bandwidth among others Such network bandwidth surpasses methods of transmission that include copper cables and microwaves. One of the technologies that is most prevalent in the optical communication sector is free-space-optical-communications, which is a result of the need for mobile communication systems. Current demands and high-capacity of free-space-optical-communications in terms of transmission capacity, dormancy, and authenticity are brought forth by rapid advancements in communication technology, massive data utilization, and optical impairments. Despite these benefits, free-space optical communications are severely impacted by harsh weather situations like mist, precipitation, blizzard, fume, soil, and drizzle debris in the atmosphere, all of which have an impact on the Quality of Service (QoS) rendered by the systems. The primary goal of this article is to optimize the QoS using the ensemble learning models Random Forest (RF), ADaBoost Regression (ADBR), Stacking Regression (SR), Gradient Boost Regression (GBR), and Multilayer Neural Network (MLNN). To accomplish the stated goal, meteorological data, visibility, wind speed, and altitude were obtained from the South Africa Weather Services archive during a ten-year period (2010–2019) at four different locations: Polokwane, Kimberley, Bloemfontein, and George. We estimated the data rate, power received, fog-induced attenuation, bit error rate and power penalty using the collected and processed data. For optical attenuation is about 1 dB/km, at a wavelength of 1550 nm across the study locations, Polokwane, Kimberley, Bloemfontein, and George, the transmitted power of the FSO link has an ability to send the data of the rate of $1.62 \times 10^{14}$, $7.68 \times 10^{12}$, $2.80 \times 10^{12}$, $and$ $2.90 \times 10^{13}$ bps correspondingly. At a wavelength of 1550 nm and a visibility of 1 km, the values of the factor of diminution across all the places, Polokwane, Kimberley, Bloemfontein, and George, are 2.13, 2.13, 2.13, and 2.13 dB/km, respectively, The RMSE and R-squared values of the model across all the study locations, Polokwane, Kimberley, Bloemfontein, and George, are 0.0073 and 0.9951, 0.0065 and 0.9998, 0.0060 and 0.9941, and 0.0032 and 0.9906, respectively. The result showed that using ensemble learning techniques in transmission modeling can significantly enhance service quality and meet customer service level agreements. Regardless of the linked climate vicissitude along the transmission connection, the outcomes demonstrated that the ensemble method was successful in efficiently optimizing the signal-to-noise ratio, which in turn enhanced the QoS at the point of reception.

1. Introduction

A Free Space Optics (FSO) communications system is an optical communication technology that transmits data without wires for networking between computers and telecommunications by using light (laser beam, visible, infrared, and ultraviolet band) propagating in a vacuum or a desultory vessel via a Line-of-Sight (LOS) as shown in figure1. (Maswikaneng et al., 2018 and Maswikaneng et al., 2023). It provides a realistic substitute for addressing the issue of network bottlenecks and in addition to enhancing traditional radio frequency/microwave communications (Sharoar et al., 2020 and Maswikaneng et al., 2022). FSO is appropriate for providing a changeable interactive response that offers a remarkably adaptable, authorized inexpensive, instantaneous, and unconstrained permission distant internet network for a variety of uses, including language, information point, footage, and diversion (Maswikaneng et al., 2018 and Maswikaneng et al., 2023). In their 2024 study, Choyon and Chowdhury examined the efficiency of FSO communication link under various modulated schemes in the midst of severe atmospheric instability. The findings indicated that BPSK required less power correction than other modulating schemes. FSO communication provides a strong supplementary answer framework for future networked communications, regardless if it is used as a blended or an independent system. It offers various benefits if used in 5G technology mobile data backing up systems, such as rapid communication, modest bit error rate (BER), adaptable access to networks, safe propagation, substantial data speed up to 10 Gbps, ease of setting up, comparatively low cost of deployment, resistance to electromagnetic interference, and without a licensed range. FSO links are easy-to-use technologies with high-end user security because they are infrared-based on LOS and independent of transmitting mechanisms, (Garlinska, et al, 2020; Kolawole et al, 2022, Maswikaneng, et al, 2022,). The homodyne recognition constructed several ray Wavelength Division Multiplexing (WDM) FSO communication system was investigated by Gupta and Goel 2024 under a variety of circumstances. The authors demonstrated that WDM and EDFA enhancement achieved a bandwidth capability of 320 gigabits per second and that the FSO link upheld a range between 1.8 and 4.3 km throughout heavy rainfall. Since the Internet and its services represent the backbone of the information technology (IT) sector, an increasing demand for high-speed Internet calls for a large expansion of broadband and capacity. Increasing information and web supplies cause bottlenecks in prevalent radio frequency (RF) technology, requiring a switch to FSO technologies (Rashid and Semakuwa 2014; Kesarwani, et al, 2018). Understanding Quality-of-Service (QoS) as a system of measurement to account for the organization's dependability and signal strength is of relevance to system designers. The environmental effect on the transmission channel is the primary challenge confronting FSO systems. QoS, a technique for evaluating the degree of efficacy of a decent link, is observed to be a challenging task since it hinges on mainly on quadruple elements: packet delivery ratio, throughput, jitter, and end-to-end latency (Dhafer and Thulfiqar 2013, Ojo et al., 2021). Due to their reliance on time and susceptibility to internal factors, these essential components have to be resolved as a whole (Sangeetha et al. 2024).

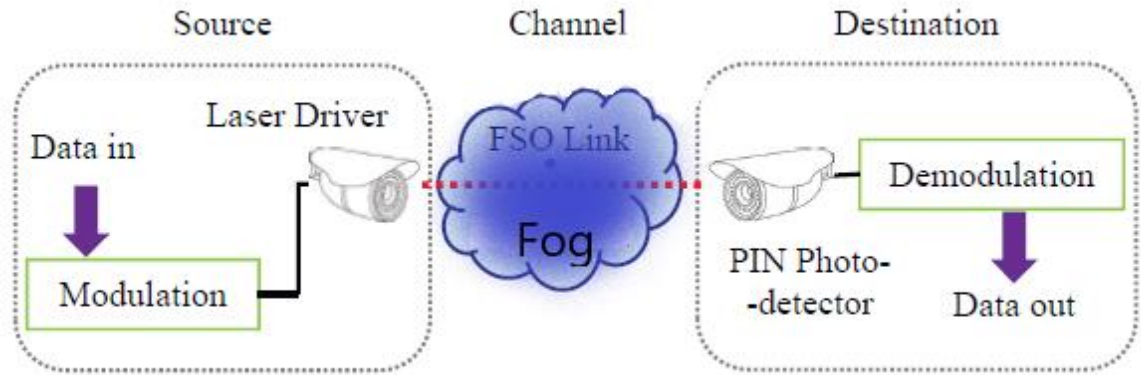

Figure.1 FSO communication systems with CSI.

Given the recent requirement for longer wavelengths and wider broadband, diminution on FSO transmission paths can therefore meaningfully distress the QoS. The emission wavelength at which the diminution problem is most apparent is below 1550 nm. Due to their reliance on time and susceptibility to internal factors, these essential components must be fixed as a unified unit. (Zaghar et al., 2013, Ojo et al., 2021). In terms of client fulfillment and system efficacy testing in connection to service level agreements (SLAs) between service providers and their corresponding clients, QoS is crucial in FSO (Lu 2002 and Maswikaneng, et al, 2023). By QoS, the capability of a structure to react strategically to thresholds related to coding, modulation, data rate, and signal power may undoubtedly guarantee the SLA's longevity and reliability (Maswikaneng, et al, 2023). The term Quality of Experience (QoE) refers to how satisfied a consumer is via an experience overall and how good they think it is, especially in multimedia and telecoms. It includes an unbiased evaluation of a consumer's experience that takes into account elements like flexibility, intuitive use, visual and aural clarity, and dependability (Panahi et al, 2024). Enhancing QoS may substantially boost QoE by managing the operational variables which influence how a service is perceived by users. In applications like playing games online, streaming videos, and actual time communication, internet service suppliers may reduce user disruptions, delays, and quality deterioration by managing QoS metrics include broadband, delay, jitter, and transmission loss. For example, smoother video playback and more rapid interactions result from lower latency and packet loss, which minimizes interruptions that lower user pleasure. Additionally, increased transmission speed translates into improved multimedia integrity and higher quality, which makes the user's experiences intriguing and pleasurable. Although the subjective character of QoE may not be entirely explained by QoS enhancements alone, they do provide a solid technological basis for improved service quality, which raises user happiness and assessed value. In this paper, the BER and SNR have been used as metric tools to determine the QoS in FSO considering foggy conditions. The reduced value of BER indicates a high value of SNR, and this implies good QoS as well as improving the QoE.

  A decline in the Signal-to-Noise ratio (SNR) due to climatic factors on the transmission connection may result in further problems. The channel used for the transmission of signals from the point of origin to the receiver is extended for FSO lines. Furthermore, Zhu et al. (2023) introduced a method for housing price prediction using stacking ensemble learning optimization.

Their findings show that the more varied the models, the more accurate the predictions utilizing the suggested D-Stacking method could be. The suggested D-Stacking technique shows exceptional possibility in decreasing the RMSE to 0.869 and 1.029 crossways multiple data set when compared to conventional Stacking ensemble learning models. The study conducted by Mienye and Sun (2022) provided a concise yet thorough review of training, spanning from its beginnings to the most current advancements in method technology. Their investigation focused on the concepts, procedures, applications, and prospects of collaborative training. Additionally, the 3 primary types of composite techniques bagging, boosting, and stacking were examined, with an emphasis on the popular composite methods, including ADaBoost, XGBoost, LightGBM, CatBoost, and random forest. As far as we are aware, no research has been done recently on using ensemble learning techniques to anticipate the QoS in FSO networks in order to address QoS anomalies in free space optical communication systems. Therefore, it is possible to predict the reliability of QoS in FSO communication connections using ensemble learning approaches. This work aims to use ensemble learning approaches to forecast the reliability of QoS in FSO networks over a subtropical climate. Moreover, confirms how the data rate, optical power received, power penalty, channel capacity, BER, and atmospheric diminution factor affect QoS dependability. The paper remaining sections of the article are arranged hence: segment II covers method; segment III covers outcomes and discussion; segment IV includes deductions and future work.

## 2. Methodology

This section provided an overview of the location's features, the type of data used, and the methods used for this investigation.

2.1 Data collection

The meteorological visibility data during a period of ten-year (January 2010–December 2019) over Polokwane, Kimberley, Bloemfontein, and George, South Africa, was retrieved from the Climate Facilities of South Africa (SAWS) archive. Three downloads of the data were made each day at eight, fourteen, and twenty hours. The air attenuation coefficient was determined by averaging the data gathered over the ten years. Utilizing the result, the data rate, optical power received, signal to noise ratio (SNR), and optical wavelength (760, 860, 960, 1260, and 1550 nm) were computed. Information on the area of study were provided based on information from the wok of Kolawole et al. (2022). The location of the data collection is depicted in Figure 2, and Table 1 displays the operational parameters that were utilized to simulate the outcomes.

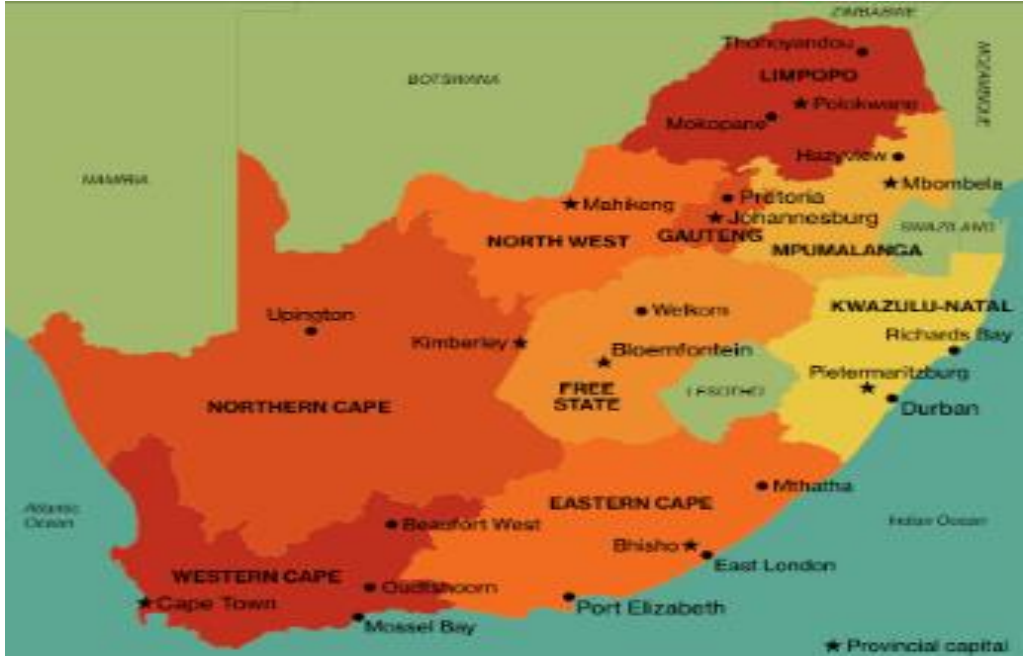

Figure2 shows the provinces where the data was collected (Kolawole et al., 2022).

Table 1: parameters for an FSO system's operation

| Operating parameters | Values |
|---|---|
| Transmitter power | 5-100 mW |
| Divergence Angle | 3 mrad |
| Source effectiveness | 4/5 |
| recipient effectiveness | 4/5 |
| wavelengths | 760,860,960,1260, and 1550 nm |
| Range | 0.1≤L≤1 km |
| Visibility | location dependent |
| Receiver Sensitivity | -40 dBm |
| Plank's constant | $6.626 \times 10^{-34}$ Js |
| PIN Load resistance | 1000 ohms |
| Boltzmann constant | 1.36E-23 J/k |
| Absolute Photodiode Temperature | 298 K |
| Dark Current | 10 nA |
| Responsivity | 0.7 A/w |
| Electrical Bandwidth | 1.0 GHz |
| Power penalty @ BER | $10^{-9}$ |

2.2.0 Simulating Optical Beam Transmission via the Atmospheric Channel

In transmitting signals in FSO communication systems, there are several impairments like atmospheric turbulence, scintillation, rain, dust, fog, snow, and back radiation that impede the quality of the signal received at the receiver. For the purpose of this research, the authors have considered the effect of fog on FSO, and this is because fog is the predominant source attenuation in the region of study. Beer-Lambert's law defines how optical signals propagate across turbulence-filled air channels. As stated in, Maswikaneng et al. (2022).

$$\Psi(\lambda, L_d) = e^{-(\Upsilon_{ext,L})} \tag{1}$$

where $\Psi(\lambda, L)$ represents the atmospheric network's transmission, $\Upsilon_{ext}$ implies the atmospheric attenuation coefficient, and $L_d$ implies transmission range. The transmittance depends on the visibility ($V$) of the weather, which indicates how transparent the atmosphere is. The visibility is the route length at which the transmittance drops to a particular propagation threshold value. (i.e., $\Psi(\lambda, L_d) = \Gamma_{th}$). The broadcast threshold, also known as the visual threshold, for optical wireless communication systems, is set at 2%, whereas for runway visual range at airports, it is set at 5% (Kolawole et al., 2017). The factor of atmospheric diminution, at $\Gamma_{th} = 2\%$ is given as

$$\Upsilon_{ext} = -\ln\frac{0.02}{V} = \frac{3.912}{V} \tag{2}$$

whereas V means visibility in kilometers. The equation in (3) can then be utilized to estimate the diminution of the air in dB.

$$\chi(L, V) = 10 \log_{10}(e)\Upsilon_{ext}(V) L \tag{3}$$

The Kruse formula can be used to determine the total extinction coefficient at wavelengths ranging from visible to near-infrared and is adapted (Kolawole et al., 2017, Ghassemlooy and Popoola, 2022, and Kim et al.,2001):

The Kruse model can be used to compute the total extinction factor from visible to near-infrared wavelength is adapted as (Kim et al., 2001; Kolawole et al., 2017; Ghassemlooy and Popoola, 2022):

$$\mu_{\mu,sct}(V) = \beta_{ae}(\lambda) = \frac{10\log_{10} T_{th}}{V(km)} \times \frac{\lambda(nm)^{-\rho_o(v)}}{\lambda_o} \tag{4}$$

where $\lambda$ is the optical signal wavelength in nanometers, $\lambda_o$ is the all-out range wavelength of the solar band and $\rho_o$ is the dimensions element delivery stricture and is written as (Kim et al.,2001):

$$\rho_o(V) = \begin{cases} 1.6, & if \quad V > 50 \text{ km} \\ 1.3, & if \quad 6 \text{ km} < V < 50 \text{ km} \\ 0.585V^{\frac{1}{3}}, & if \quad 0 \text{ km} < V < 6 \text{ km} \end{cases} \tag{5}$$

A number of investigations have computed the aerosol scattering coefficient using the Kruse model. Kim et al. (2000) discovered, however, that in cloudy conditions, the distribution of particle sizes variable in the Kruse technique is unable to reliably estimate the values of the scattered diminution factor for a range of less than six kilometers. Consequently, to correctly estimate β_ae (λ) further, (Kim et al., 2000) proposed a novel modification for the particle size related coefficient, which is as follows (Kim et al., 2000):

$$\rho_o(V) = \begin{cases} 1.6, & if \quad V > 50 \ km \\ 1.3, & if \ 6 < V < 50 \ km \\ 0.16V + 0.34, & if \ 1 < V < 6 \ km \\ V - 0.5, & if \ 0.5 < V < 1 \ km \\ 0, & if \quad V < 0.5 \ km \end{cases}$$

(6)

2.2.1 Signal to Noise Ratio (SNR)

This is a crucial communication link metric for assessing quality, as it has an inverse relationship to attenuation. SNR mathematical expression is defined in (Yasir et al. 2018 and Adebusola et al. 2024):

$Q_{snr} = P_{tmt} - 30 - 10log(G_{tmt}) + 10\log(G_{rvr}) - 20\log(\frac{4\pi}{\lambda}) - 10\log(B_{dth}T_{amb}K_{blt}) - \tau - NR_f - F_{mgn}$ (7)

where $P_{tmt}$ is the transmitted power, $G_{tmt}$ is the transmitted antenna gain, $G_{rvr}$ the receiver antenna gain, $\lambda$ wavelength, $K_{blt}$ is the Boltzmann's constant (1.38*10^34 J/K), $B_{dth}$ is the receiver bandwidth (BW = 1 MHz), $T_{amb}$ the ambient temperature in K, $\tau$ is the total attenuation in dB/km, $NR_f$ is the receiver noise figure and $F_{mgn}$ is the fade margin.

2. 2.2 optical power received and beam divergence for the FSO system

Received optical power is defined as the amount of optical that is detected by the receiver at the receiving end of the FSO link. It a crucial parameter that determines the performance and reliability of the FSO communication system. The received optical is typically measured as the input of the receiver and is a key factor in determining the quality of the received signal. It is usually measured in dBm. A higher received optical power indicates a stronger signal at the receiver, which can result in improved signal quality and increased system performance (Ghoname et al.2016). Laser dispersion is a crucial component of FSO telecommunication that affects the network's practical distance, particularly in harsh environments. The term laser diverging describes how a laser light spreads out over a range, increasing in dimension as it moves. It has an enormous effect on the signal intensity since a broader laser lowers the power efficiency at the point of reception, which lowers the standard of the signal that is transmitted (Ghassemlooy et al, 2019,). optical power received can be estimated using equation (8) (Ghoname et al., 2016 and Maswikaneng et al., 2023).

$P_{rx} = P_{tx}\frac{D^2_r}{\theta^2_{div}L^2}e^{-\frac{\beta_{ae}(\lambda)L}{10}}\zeta_{trans}\zeta_{rec}$ (8)

where $P_{tx}$ is the transmitted power, $L$ is the link distance, $D_r$ is the receiver diameter, $\theta_{div}$, is the divergence angle, $\beta_{ae}(\lambda)$, is the total attenuation factor(dB/km), $\zeta_{trans}$ and $\zeta_{rec}$ are the transmitter and receiver optical efficiency respectively.

### 2.2.4 Data rate

Ali. 2014, Stated that, for provided power transmitted laser, with divergence transmitted of $\theta$, diameter of receiver D, transmit and optical receiver effeciency $\zeta_{trans}$, $\zeta_{rec}$, the achievable data rate $R_{data}$ can be determined from (Ali 2014 and Maswikaneng et al., 2023):

$$R_{data} = \frac{P_{trans}\tau_{trans}\tau_{receiv}10^{-\gamma*\frac{L}{10}}D^2}{\pi(\theta/2)^2 L^2 E_P N_P} \qquad (9)$$

Equation (9) may be written as:

$$R_{data} = \frac{4}{\pi E_p N_b} * P_{receiv} \qquad (10)$$

where $E_p = \frac{hc}{\lambda}$ (11)

$E_p$ represents photon energy

while for optical power received in FSO can be explained as:

$$P_{receive} = P_{trans} x \frac{d_{receiv}^2}{((d_{trans}+\theta xL)^2)} x\ 10^{-\gamma L/10} \qquad (12)$$

where $P_{receive}$ received power, $\gamma$ attenuation coefficient, L length of opical length, $\theta$ is divergence angle and $d_{receiv}, d_{trans}$ represents receiver and transmitter diameters respectively.

### 2.2.5 Channel capacity

Fog has a significant impact on a FSO transmission link's channel capacity because it weakens what is transmitted through transmission loss. it is the highest amount of data that can be sent in a given amount of time, and also it is an additional metric that is assessed here in order to describe the efficiency of an FSO network. The Shannon-Hartley theorem, which provides the highest possible speed of data depending on the Signal-to-Noise Ratio (SNR) at the receiving end, is commonly used to simulate the transmission capability amid fog (Xu et al 2021). This capability for an FSO system can be written as follows:

$$\mathbb{C} = B_{dth} \log_2(1 + Q_{snr}) \qquad (13)$$

Where $\mathbb{C}$ is the channel capacity.

### 2.3.0 BER Investigation for FSO Modulation Methods

There have been several strategies proposed to attain and enhance bandwidth efficacy. (Maswikaneng et al, 2020 and Maswikaneng et al, 2023). Using the right modulation technique is one approaches to accomplish the aforementioned, and BER can be used to measure its performance. An optical transmission technology relies heavily on BER. According to (Vladimir et al. 2019), the link's strength and quantity are highly dependent on weather factors like temperature, dust, fog, and rain. The achieved BER defines the communication situation. In FSO transmitting techniques, the tenacity of an optical source is managed to send signals across the channel. A range of modulation techniques could be used to alter FSO designs in order to attain the desired BER, according to the FSO power received (Maswikaneng et al, 2023 and Xu, et al, 2021). FSO optimize the peak-to-average power balance by using an energy-saving modulation method. Amplitude modulation, which employs direct detection, is among the most basic and widely used modulation technique (Hameed, et al, 2024). Power efficiency, bandwidth efficiency, reduced cost of execution, ease of design, and tolerance to interfering ambient electromagnetic waves are all important considerations when selecting a modulation scheme (Sarieddeen,et al 2021). The next paragraphs discuss the selected modulation scheme to set the scene for the present article.

2.3.1 ON-OFF Keying (OOK)

Due to its insensitivity to ambient modulation and laser irregularities, the On-Off keying (OOK) approach is the most fundamental modulating method utilized in commercialized earthly FSO communication networks. Bits 0 and 1 are transferred in this modulated method. Return to Zero (RZ) or Non-Return to Zero (NRZ) pulses can be used alongside the modulated approach.

In NRZ-OOK, the maximum power's optical wave represents a binary code "0," but the laser wave of maximum power transfer represents a binary code "1." There is an alternative between 0 and 1 for the optical supply attenuation rate "ae." The emblem's period is equal to the optical wave's limited time span. The likelihood of a photoreceptor failure affecting NRZ-OOK-coded image information can be expressed as a measurement of the SNR as follows: In NRZ-OOK, the color pulse of peak power represents a binary code "0," but the laser wave of maximal power transfer represents a binary code "1." The optical origin attenuation rate might be between 0 and 1. "T" is the limiting time frame of the optical wave.

$$BER_{NRZ-OOK} = \frac{1}{2} erfc \left[ \frac{1}{2\sqrt{2}} \sqrt{SNR} \right] \tag{14}$$

The bit rate needed for the NRZ is equal to the bandwidth ($B_{req} = R_b$").
At the expense of increasing the information transmission capability, the necessary SNR in RZ-OOK is equivalent to 1/2 (-3dB) of the SNR required for the standard NRZ-OOK for the proportionate performance of BERs (Maswikaneng et al., 2020). The BER of the RZ-OOK may be shown as an element of Signal to Noise Ratio (SNR):

$$BER_{RZ-OOK} \frac{1}{2} erfc \left[ \frac{1}{2} \sqrt{SNR} \right] \tag{15}$$

RZ-OOK has a shorter NR-OOK oscillating period compared to NRZ-OOK, which improves the performance of power at the cost of a larger communication bandwidth (Maswikaneng et al., 2020).

## 2.4.0 Ensemble learning techniques

An ensemble technique is a kind of artificial intelligence that raises an activity's estimated efficiency by combining the estimates from multiple estimation approaches or algorithms. In this way, an ensemble technique obtains a meta-estimator (Mienye. and Sun, 2022). The model elements that comprise an ensemble are basic learners, or basic estimation tools. Ensemble techniques, which leverage the strength of a plurality of knowledge sources, are based on the notion that the opinions of a group are more influential than those of any individual team member. Combined techniques are widely used in many application fields, including financial and commercial analytics, healthcare, safety concerns, education, production, system optimization, leisure activities, and many more (Breiman, 2001; Ahrens 2022). The trade-off between precision and complexity, often known as bias-variance, which numerous artificial intelligence approaches encounter affects how well they adapt to new data. To overcome this trade-off, ensemble techniques employ several element models. To be efficient, an ensemble needs two things in particular: (1) variety within the ensemble and (2) model pooling for the ultimate forecast. figure 3 depicts the general block diagram for Ensemble learning technique.

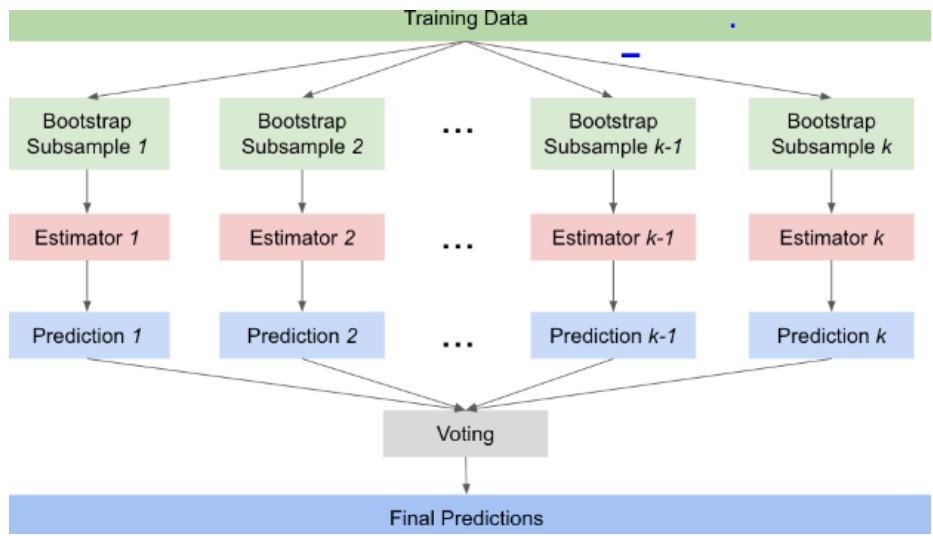

Figure.3; Architecture of Ensemble learning technique (Vijayaganth, et al,2023)

## 2.4.1 Random Forest (RF)

Several tree-based techniques depend on the fundamental idea of predicting the dependent variable by dividing the learning sample into portions according to threshold values of a chosen covariate, with each split trying to optimize the algorithm's fit. For separate tree $T_\alpha \in \{T_\alpha\}_1^\alpha$ of $\alpha$ these trees, RF attracts with extra since the learning specimen a bootstrap specimen, $\mathbb{z}^\alpha = \{\mathbb{z}_1^\alpha, \ldots, \mathbb{z}_m^\alpha\}$ of magnitude m, wherever $\mathbb{z}_1^\alpha = (\psi_i^\alpha, \chi_i^\alpha)$. Next, employing an acceptable threshold for the covariate x b I that yields the smallest total of square error values, the theorem divides the bootstrap specimen into two segments. It thus continues dividing the specimen till a segment that has the fewest occurrences is reached. The smallest possible size of a segment is usually set at a single measurement for regression and five observations for classification (Breiman, 2001; Ahrens 2022). To prevent excessive fitting, the value of this variable is capable of being adjusted. Upon repeating the procedure for the remaining B- trees, the forecasts are aggregated using either an overwhelming majority for classification or an average in each instance of regression (Breiman, 2001; Ahrens 2022).

$$\mathcal{F}^{\propto}(\chi) = \begin{cases} \frac{1}{\propto}\sum_{\propto-1}^{\propto} T_{\propto}(\chi) \ \text{if regression} \\ \text{mode } \{C_{\propto}(\chi)\}_1^{\propto} \ \text{if classification} \end{cases} \quad (16)$$

We have discussed bagging as it relates to tree-based estimation tools thus far. The fact that the bulk of the trees' splits will probably be chosen from the same stronger predictors is one problem with this. This could result in a strong relationship among the trees, that would not improve the model's exactness. The dilemma is spoken by choosing an arbitrary subsection of n out of a covariate to be studied for a separation, where n is alternative factor in conjunction with $\propto$ and the least segment scope to be adjusted with the mutual selections for n = l/ 3 for a decrease and m = $\sqrt{k}$ for a categorization. Moreover, take note of the observational prediction inaccuracy j is determined through linking its actual worth of $\psi_j$ to the aggregate forecast $\mathcal{F}^{\propto}$ this $\mathbb{z}_1^{\propto} \notin \mathbb{z}^{\propto} \forall_{\propto}$. Out-of-Bag (OOB) loss prediction is a RF feature that approximates n-fold cross-validation loss estimate as $\propto \to \infty$ (Mienye, and Sun, 2022).

## 2.4. 2 STACKING REGRESSION (SR)

A method called stacking is used to put together regression or classification theorems adopting two layers predictor. This 1$^{st}$ step includes all of the control theorems that are adopted to forecast the consequences of the simulated samples. The meta-regressor, that creates new predictions based on all of the underlying models' forecasts as inputs, makes up the second layer (Vijayaganth et al; 2023). Typically, the meta-estimator is a chaotic model that combines the base estimators' estimates chaotically. This extra complexity comes at the cost of stacking frequently overfits, especially in the case of turbulent situations. To avoid overfitting, stacking is sometimes used with k-fold cross-validation to make sure that distinct baseline predictors have not been trained on the same data. This lessens the possibility of overfitting while regularly boosting ability and variation (Mienye, and Sun, 2022). Figure 4 depicts the stacked regressor model's structure. To prevent overfitting and ensure that the validation sample is clean, cross-validation is employed instead of learning the foundation trainers and meta-model on the identical sample. In actuality, stacking is a more complex form of cross-validation (Ahrens et al., 2022). Explicitly, we begin with a collection of data. $Q = (X_j, \theta_j)_{j=1}^m$ ("level 0" training specimen), where $X_j$, is the reply variables and $\theta_j$ is a predictor vector. First randomly split all the observations into H equal-sized folds $Q_1, \ldots, Q_H$. Define $Q_H$ and $Q^{(-h)} = Q - Q_h$ to be the validation and training samples for the $h^{th}$ fold of the H-fold cross-validation. Consider that we have a set of base learners $P_l^{(-h)}, l = 1, \ldots, L$ persuaded by L estimators or (algorithms), which are also called level 0 generalizers, for each training sample $Q^{(-h)}$. To complete the training process, the final base learners $P_l, l = 1,..,L$ are derived using all the observations in $Q$. The base learner for each algorithm is decided by

$$\widetilde{F}_l(\theta_j) = arg^{min}_{\widetilde{F}_l(\theta_j)} \sum_{j=1}^{M}[X_{j,} - \widetilde{F}_l(\theta_j)]^2, for\ l = 1,...L. \tag{17}$$

Where $\widetilde{F}_l(\theta_j)$ denotes the prediction of model $P_l$ on $\theta_j$. Now we have L predictions for each response variable $X_{j,}$ which are used as inputs of the new "level 1" learning sample.

$$Q_{CV} = \left((X_{j,} \widetilde{F}_1(\theta_j), ..., \widetilde{F}_l(\theta_j)\right). \tag{18}$$

Traditionally, the level 1 sample will be used to select one base learner with the best performance, i.e $\widetilde{F}_l(\theta_j)$ that minimizes $\sum \left(X_{j,} - \widetilde{F}_l(\theta_j)\right)^2$. The stacking idea is to use some algorithms (or "level 1" generalizer) to derive the meta–model $\hat{P}$ for $X_{j,}$ as a function of the L predictive values. Consider a linear regression where the stacking estimate of the weight $\widehat{U}_t$ is simply obtained by minimizing the sum of squared residual of the linear regression of $X_{j,}$ on $\widetilde{F}_1(\theta_j), ....., \widetilde{F}_l(\theta_j)$. In particular, (Breiman, 1996; Ahrens et al, 2022).

$$\widehat{U^t} = arg^{min}_{U} \sum_{j=1}^{m}[X_{j,} - \sum_{l=1}^{L} U_l \widetilde{F}_l(\theta_j)]^2 \tag{19}$$

The concluding forecaster of the meta-model for $X_{j,}$ is $\sum_l \widetilde{U}_l^t \widehat{F_l}(\theta_j)$. To guarantee that the predictive accuracy of layering regression would surpass that of the single best base learner, all the elements in the vector $\widehat{U^t}$ are restricted to be non-negative and to sum to one (Ting and Witten, 1999). Suppose there is no restraint on the weights and $F_l(\theta_j)$ are highly correlated, the $U_l$ are obtained using least squares. Then it might happen that the final predictor $\sum_l U_l^t F(\theta_j)_l$ will stay away from the range $[min_l F_l(\theta_j), max_l F_l(\theta_j)]$ and generalization to the population can be poor (Breiman, 1996; Ahrens et al, 2022 Recall that in a unique scenario when we limit the girth tensor U^t to include a single component and the remainder is zilch, it essentially transforms into a V-fold cross-validation, wherein the ultimate algorithm is determined by selecting the base learner with the highest prediction accuracy. Stacking is more versatile since it allows us to mix the models in (19) using any approach other than linear regression. The girths could also be subject to the input location $\theta$ (Friedman, 2017).

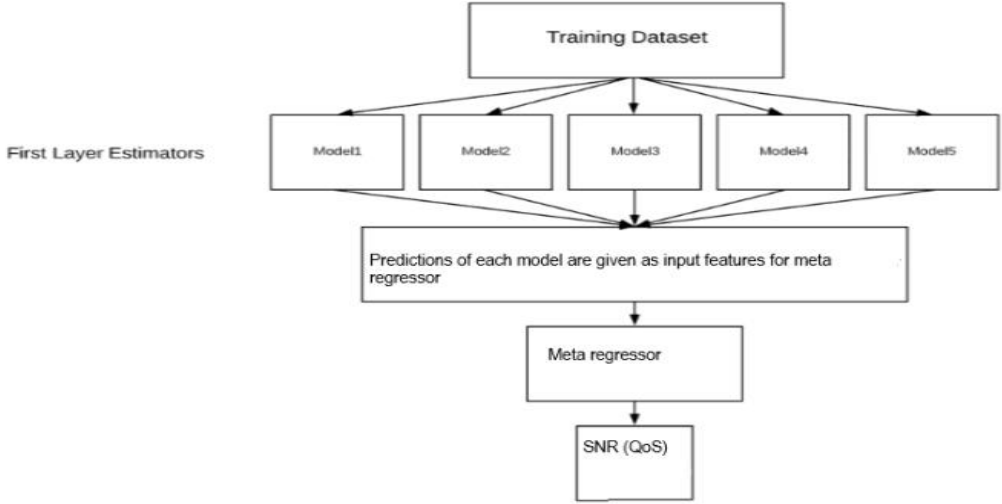

Figure 4 Structure of stacking algorithm:

2.4.3 AdaBoost Algorithm (ADBR)

The AdaBoost theorem is an enhanced allocation theorem that can transform a collection of feeble predictors into a strong predictor. These techniques often start by training a base predictor from the original training specimen using a base identify technique, and its prediction performance is merely marginally better than arbitrary guesswork. Those mistakenly classified specimens are then given extra weight when the specimen weight is adjusted based on the base predictor's output. and retrain the subsequent foundation learner using the modified specimens. The ultimate predictor is formed by adding weights to these base learners throughout the iterations. The explanation of the AdaBoost method follows (Mienye and Sun, 2022).

Let $\mathbb{S} = \{(\chi_1, \psi_1), \cdots, (\chi_j, \psi_j), \cdots, (\chi_m, \psi_m)\}$ imply the learning specimen scheduled in the double sorting condition. $\chi_j \in \Psi \subseteq P^m$ is the $j^{th}$ occurrence. $\psi_j \in \Psi = \{-1, +1\}$ is the degree characterization connected through $\chi_j$. ADBR theorem is built on the add-on techniques, that is the direct amalgamation of the base predictors $\xi_k(\chi)$:

$$\mathcal{F}(\chi) = \sum_k^K \delta_k \xi_k(\chi) \tag{20}$$

where k = {1 …, K} connote the reiteration quantity, $\xi_k(\chi)$ are the base trained predictors after the foundation sorting theorems whose predicted capability is fair improved than arbitrary guess work and $\delta_k$ are the girth factors. In (20), $\xi_k(\chi)$, are learned after the foundation predication theo rem built on the build-up specimen through the allotment $E_k$ at k reiteration. The outline of $E_k$ is a standardized girth course.

$$\xi_k(\chi) : X \rightarrow (-1, +1). \tag{21}$$

The allotment $E_k$ symbolizes the girth of respective occurrence in $\mathbb{S}$ at k reiteration. $E_1$ can be computed like

$$E_1(j) = \frac{1}{m}, \quad j = 1,2, \dots, m. \quad \text{and}$$

$E_{k+1}$ are computed like

$$E_{k+1}(j) = \frac{E_k(j)}{\mathbb{Z}_k} e^{(-\delta_k \psi_k \xi_k(\chi_j))} \quad j = 1, 2, ..., m \tag{22}$$

Where $\mathbb{Z}_k$ are the normalized factors and are estimates as

$$\mathbb{Z}_k = \sum_{j=1}^{m} E_k(j) e^{(-\delta_k \psi_k \xi_k(\chi_j))} \tag{23}$$

From (22) and (23) we identify so $E_{k+1}$ are modified after $E_k$. Accordingly the specimen that are predicted erroneously in $\xi_k(\chi)$ maybe heavier girths in k+1 reiteration. Assumed buildup set $\mathbb{S}$ and specimen girth $E_k$, the entity $\xi_k(\chi)$ is to minimize the regression error $\varepsilon_k$. $\varepsilon_k$ is computed as

$$\varepsilon_k = P[\xi_k(\chi_j) \neq \psi_j] = \sum_{j=1}^{m} E_k(j) I[\xi_k(\chi_j) \neq \psi_j]. \tag{24}$$

where P [·] signify the likelihood and I [·] signify the logical values

In eq. (20), $\delta_k$ determines the significance of $\xi_k(\chi)$ in the concluding predictor and are computed as follows:

$$\delta_k = \frac{1}{2} \ln\left(\frac{1-\varepsilon_k}{\varepsilon_k}\right) \tag{25}$$

From Eqns. (24) and (25) we could identify that once $\varepsilon_k < 0.5$, $\delta_k > 0$. And $\delta_k$ might upsurge through $\varepsilon_k$ decline. absolutely, in this technique, ADBR lowers the ascending cost coefficient. (wang et al, 2019).

It is also evident that $\varepsilon_k$ in the ADBR theorem needs to be less than 0.5. Actually, the rate of error of a dual predictor is constantly not higher than 0.5. For binary classification problems, if the rate of error $\varepsilon$ of feeble predictor:

$$\xi(\chi) = \begin{cases} 1, & \chi \in \Omega \\ -1, & \chi \notin \Omega \end{cases} \tag{26}$$

is greater than 0.5, the classifier can be applied.,

$$\xi_c(\chi) = \begin{cases} -1, & \chi \in \Omega \\ 1, & \chi \notin \Omega \end{cases} \tag{27}$$

to substitute $\xi(\chi)$, that causes the rate of error $\varepsilon_c$ convert to $1-\varepsilon$. At that point, the $\varepsilon_c$ is < 0.5. This shows that $\varepsilon_k$ might constantly be < 0.5 if equivalent to 0.5. However, it is highly unlikely that a regressor's error rate will coincide with 0.5. Afterwards unremitting reiteration, the concluding regressor is

$$\mathcal{F}(\chi) = sign(f(\chi)) = sign(\sum_{k=1}^{K} \delta_k \xi_k(\chi)) \tag{28}$$

2.5.4 Gradient Boosting (GBR)

Gradient Boosting is an additional ensemble approach that utilizes the tree-based models (Friedman, 2002). Gradient boosting targets to minimize bias by beginning with an underfit model and improving the residuals with each subsequent tree n ∈ 1..., N until it reaches its ultimate form, in contrast to random forest, which focuses on variance reduction as given by (Ahrens et al, 2022).

$$\psi(\theta) = \psi_{o\,(\theta)} + \gamma \sum_{n=j}^{N} \psi_n(\theta) \qquad (29)$$

where $\gamma \in (0,1)$ is the rate of learning. To prevent overfitting, this hyperparameter can be adjusted to regulate how each of the M trees affects the final prediction. Although it takes longer to train than random forest, this approach frequently yields estimates that are more accurate. The first step in using gradient boosting to solve a regression problem is calculating the first term of 29: (Ahrens et al, 2022).

$$\psi_{o\,(\theta)} = arg_\beta^{min} \sum_{j=1}^{T} D(x_j, \beta) \qquad (30)$$

where $D(x_j, \beta)$ is a loss function and γ are the predictions, each of which corresponds to one of I terminal nodes of tree m given some predictive rule $\psi(\theta) = \beta_j$. Although there are multiple loss functions available, squared error loss, defined as $D\left(x_j, \psi(\theta_j)\right) = \frac{1}{2}\left[x_j - \psi(\theta)_j\right]^2$, is suitable for discrepancy and, thus, is a general choice. The subsequent terms 1 through N are subsequently reached by the algorithm. In this section, the artificial leftovers provided by the reduction function's inverse gradient are calculated with the forecasts that were assessed at the preceding tree's forecasts., i.e

$$\lambda_{j,n} = -\left[\frac{\partial D(x_j,\beta)}{\partial \psi(\theta)_j}\right]_{\psi_n\,(\theta)=\psi_{n-1}(\theta)_j} \forall_i \qquad (31)$$

which are then used as respective targets for each *I* terminal nodes to fit the tree. Finally, it is calculated as:

$$\psi_n(\theta) = \psi_n - 1(\theta)_j + \sum_{i=1}^{I} \beta_{i,n} I(\theta \in R_{i,n}) \qquad (32)$$

where $\beta_{i,n} = arg_\beta^{min} \sum_{\theta_j \in R_{i,n}} D(x_j, \psi_{n-1}(\theta)_j + \gamma) \forall_j \qquad (33)$

and $\sum_{i=1}^{I} \beta_{i,n} I(\theta \in R_{i,n}) = \Gamma_n(\theta)$ is a formal expression for tree m with I terminal nods denoted by $R_{i,n}$. The prediction of the preceding tree is taken into consideration in the loss function minimized by each of the predictions $\beta_{i,n}$ that differs from the one in (29) alone. Furthermore, a classification problem can also be solved via gradient boosting. Similar to the previous description, the process involves computing each of $\psi_n(\theta)$ K times, where K is the number of classes, and using negative log-likelihood as the loss function (Ahrens et al, 2022).

2.4.5 Artificial Neural Network Application (ANN)

An enormous number of little processing units known as nerve cells provide the foundation of complex net structures known as artificial neural networks (ANNs). In real-world applications, backpropagation (BP)-trained ANN are the most widely used techniques. (Chen et al., 2019). The BP method compares the goal and output values and minimizes model error by using the gradient steepest descent strategy. Figure 5 depicts a typical ANN construction.

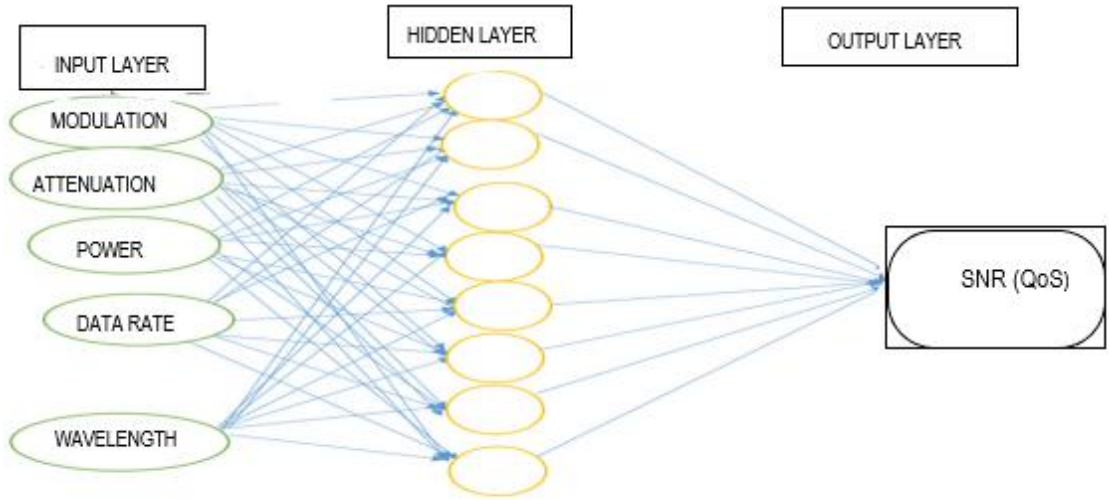

Figure5: Artificial Neural Network Layers

In wireless transmission, convectional classical statistical solutions have been overtaken in recent years by neural networks. The working concepts of the human brain are similar to this model. An ANN model makes use of the nodes that transmit signals with knowledge through a structure in place of brain neurons. An artificial neural network is based on opinion. It takes the input signal $\omega_j$ with weight $\omega_j$ and bias $b_j$ adds them through a summation function and then passes them through an activation function $\mathcal{F}$ to give the ANN algorithm its overall nonlinear properties and enable learning of almost any function as shown in Fig4. After that, the perceptron transfer coefficient can be written as (Cifuentes et al, 2020):

$$y = \mathcal{F}\left(\sum_{j=1}^{m} \omega_{kj} x_j + b\right) \tag{34}$$

The input layers receive an input vector and relays its findings to every neuron in the concealed layers. Layers are connected by networks weights. The weight and bias factors are modified throughout the training stage. The period of training terminates whenever the goal and the output have the same value, or whenever the training stage accomplishes its designated repetition result (Saif et al. 2020). Unlike other approaches, the ANN technique enters the execution stage promptly following retraining. The developed ANN model is challenging to comprehend, though. Furthermore, when the total amount of layers and/or neurons is insufficient to represent the complexity of the work to be learned, the optimization process may fail while retraining to achieve an adequate minimum quantity. In the expected ANN structure, the weighted input signal from the hidden layers is obtained along with a bias and represents the values of the predicted variable through a single node in the output layer SNR and five input nodes in the input layer (modulation, data rate, attenuation, power, and wavelength). This study's multi-layer perceptron model serves as its framework, and the input-output mapping is described by (Lionis, 2021):

$$y = \mathcal{T}\left(\sum_{i=1}^{n_1} \omega_j \mathcal{F}\left(\sum_{j=1}^{m} \omega_{kj} \mathfrak{x}_j + b_k\right) + c\right) \tag{35}$$

where $\mathcal{T}$ and $c$ denote the activation function and bias for the output layer, correspondingly. Given this justification, more intricate systems with several hidden levels might be taken into consideration. In the course of learning, which is generally referred to as training, the weights vector W, which describes the asymmetric visualization, is defined to correspond to the intended outputs while reducing a specified loss rate. An activation operation, usually expressed by one of the subsequent operations, defines each of the n hidden neurons.

The Hyperbolic function of the Tangent: $\mathcal{F}(t) = \frac{e^t - e^{-t}}{e^t - e^{-t}}$ (36)

Sigmoid function: $\mathcal{F}(t) = \frac{1}{1+e^{-t}}$ (37)

Remedied Lined Unit (ReLU) function : $\mathcal{F}(t) = max(0, t)$ (38)

Gaussian function: $\mathcal{F}(t) = e^{-t^2}$ (39)

Direct Function $\mathcal{F}(t) = t$ (40)

In order to visualize the vast climate data collected at the meteorological location, a two-layered feed-forward containing sigmoid hidden neurons and linear output neurons was trained for a greater amount of nodes using conventional Levenberg-Marquardt training. All of the ANN calculations were carried out using MATLAB's neural net matching tool. The testing, training, and validation portions of the full set of information are divided three times using the ANN system. While learning, the neural network is shown a portion of the training set, and it is modified based on its error. The algorithm's adaptation is assessed using the validation subset, and training is terminated once adaptation ceases to improve. The testing subset assesses the network's effectiveness following training. Thirty percent of the total data set was used for network training, fifteen percent was used for validation, and thirty percent was used to evaluate the performance of the network. (Maswikaneng et al., 2023).

Table2: Evaluation metrics (RMSE, MAPE, MAE and $R^2$)

| Meaning | Abbreviation | Explanation | Formula |
|---|---|---|---|
| Mean absolute percentage error | MAPE | Lesser values are improved | $\frac{1}{n}\sum_{i=1}^{n}\left|\frac{Q_{as} - Q_{ps}}{Q_{as}}\right|$ |
| Mean absolute error | MAE | Lesser values are improved | $\frac{1}{n}\sum_{i=1}^{n}|Q_{as} - Q_{ps}|$ |
| Root mean square error | RMSE | Lesser values are improved | $\sqrt{\frac{\sum_{i=1}^{n}(Q_{as} - Qs_{ps})^2}{n}}$ |

| Mean square error | MSE | Lesser values are improved | $\frac{\sum_{i=1}^{n}(Q_{as}-Q_{ps})^2}{n}$ |
| --- | --- | --- | --- |
| Coefficient of Determination | $R^2$ | Values closer to one are better | $1 - \frac{\sum_{i=1}^{n}(Q_{as}-Q_{ps})^2}{\sum_{i=1}^{n}(Q_{as}-Qm_{ps})^2}$ |

where i is the number of terms in the data table and n denotes the number of samples. The actual and expected values of the signal-to-noise ratio are also denoted by $Q_{as}$, $Q_{ps}$ and $Qm_{ps}$ implies the mean.

3.0  Results and Discussion

3.1.1 The impact of vision attenuation coefficients of fog at various optical communication frequencies.

Figures 6(a-e) show the fluctuation of visibility on fog attenuation throughout a variety of optical communication space wavelengths, which are frequently used in free-space optical communication systems, as accomplished using the upended Kruse model. The results show that the factor of diminution decreases as visibility increases while also decreasing as wavelength increases throughout all of the studied areas. This is the result of an inverse relationship that occurs between the factors of diminution and visibility, as described in equation (4). For an instance, at a wavelength of 1550 nm and a visibility of 1 km, the values of the factor of diminution across all the places, Polokwane, Kimberley, Bloemfontein, and George, are 2.13, 2.13, 2.13, and 2.13 dB/km, respectively, whereas at the same visibility with a wavelength of 760 nm, the values of the factor of diminution across the same locations are 3.240, 3.010, 3.236, and 2.923 dB/km, in that order. The percentage of declination while broadcasting an FSO signal from 760 nm to 1550 nm at 1km in Polokwane, Kimberley, Bloemfontein, and George is 34.26%, 29.24%, 34.18%, and 27.12%, respectively. The results suggest that little of an FSO signal is diminished when sent at a higher wavelength. Thus, telecoms service providers should use higher wavelengths for better connection efficacy and QoS.

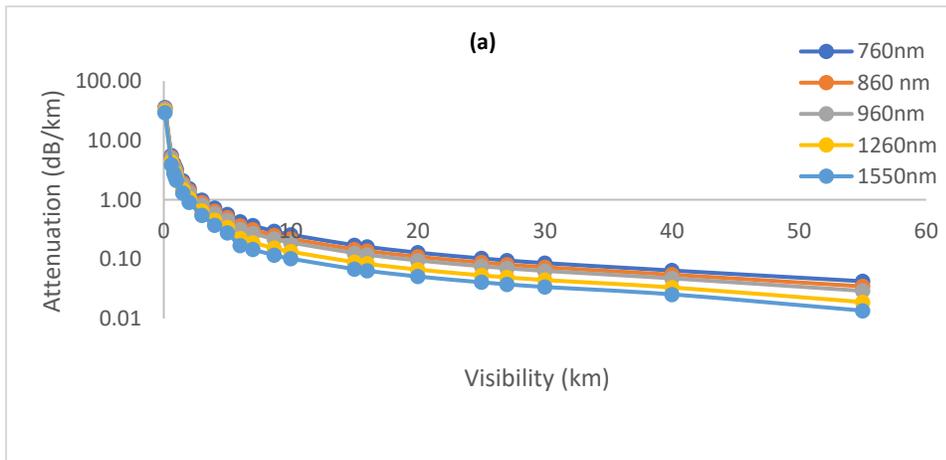

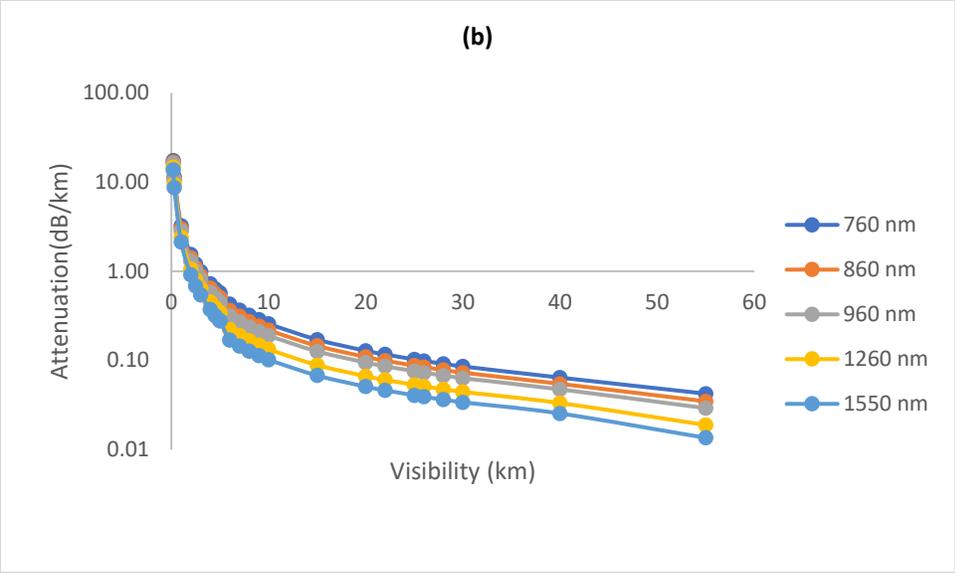

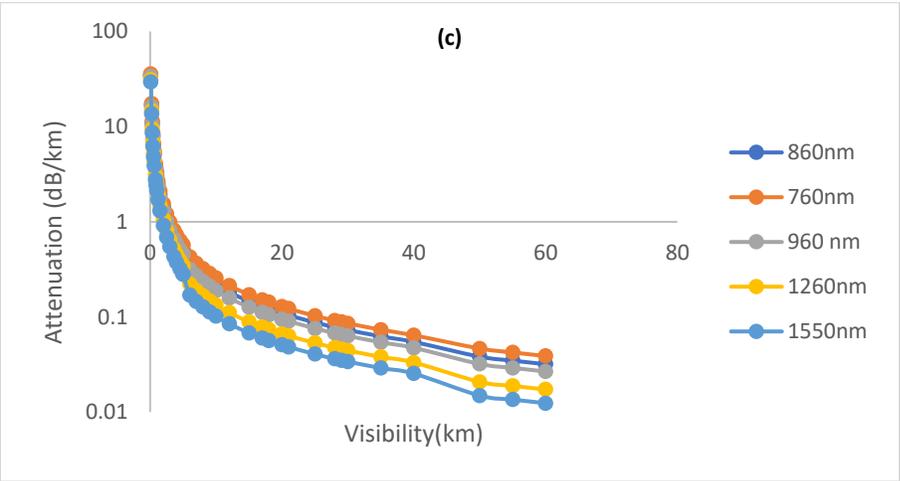

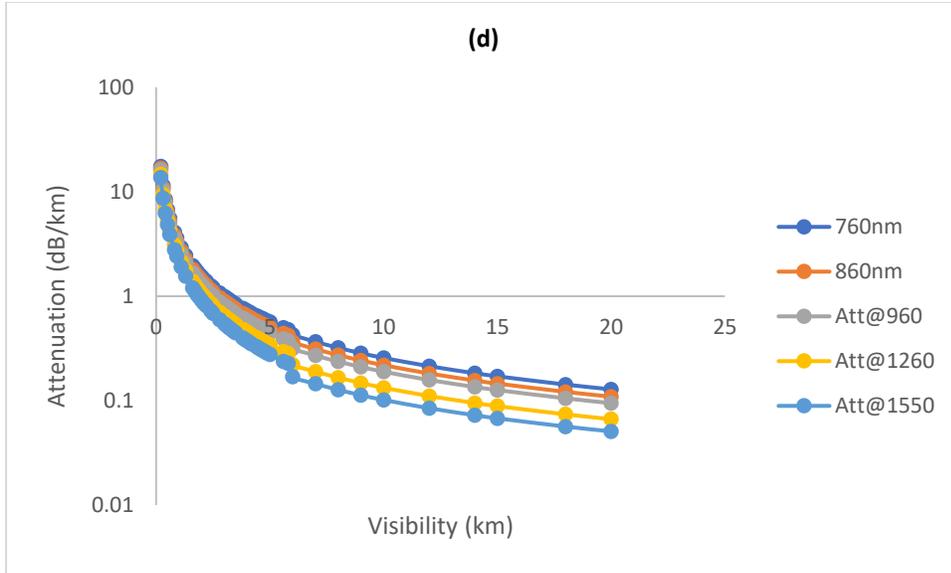

Figures 6. The distributions in variations of attenuation coefficient and visibility at different wavelengths for (a) Polokwane (b) Kimberley (c) Bloemfontein and (d) George.

3.1.2 Evaluation of data rate and specific attenuation

A higher amount of received power leads to lower specific attenuation, a higher data rate and a higher link margin. Higher attenuation will lower the data rate, as shown in Figures 7(a-d), across the studied locations. When optical attenuation is about 1 dB/km, at a wavelength of 1550 nm across the study locations, Polokwane, Kimberley, Bloemfontein, and George, the transmitted power of the FSO link has an ability to send the data of the rate of $1.62 \times 10^{14}$, $7.68 \times 10^{12}$, $2.80 \times 10^{12}$, and $2.92 \times 10^{13}$ bps accordingly, while at a 750 nm value of wavelength and at the same value of specific attenuation, the achievable data rate values across the same locations are $5.673 \times 10^{13}$, $2.63 \times 10^{12}$, $9.51 \times 10^{11}$, and $1.25 \times 10^{13}$ bps accordingly. The percentage increase in transmitting signal for high throughput that would intend to improve the QoS across the locations Polokwane, Kimberley, Bloemfontein, and George of the study is 65.00%, 65.76%, 66.04%, and 57.19%. Also, by increasing the optical transmitted wavelength, both under clear sky conditions or under foggy conditions, the data rate can be improved, as indicated in Figures 7(a-d).

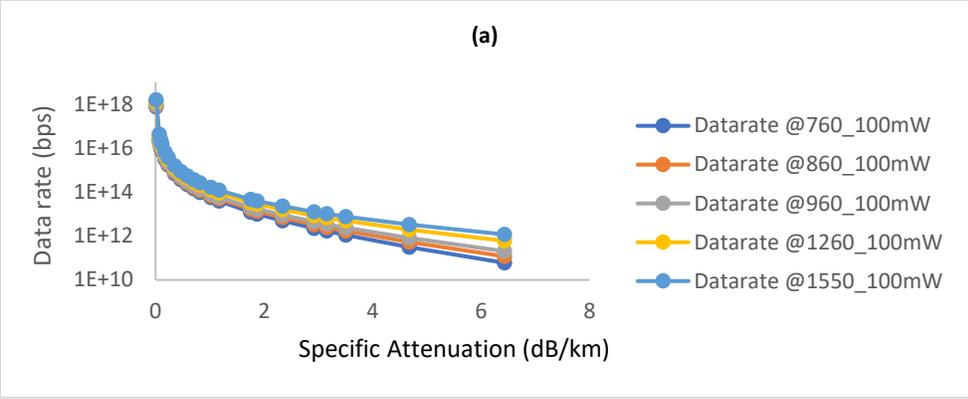

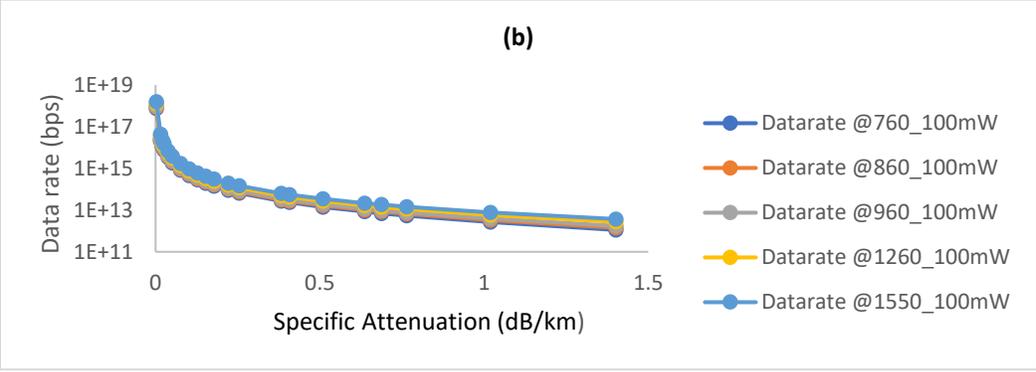

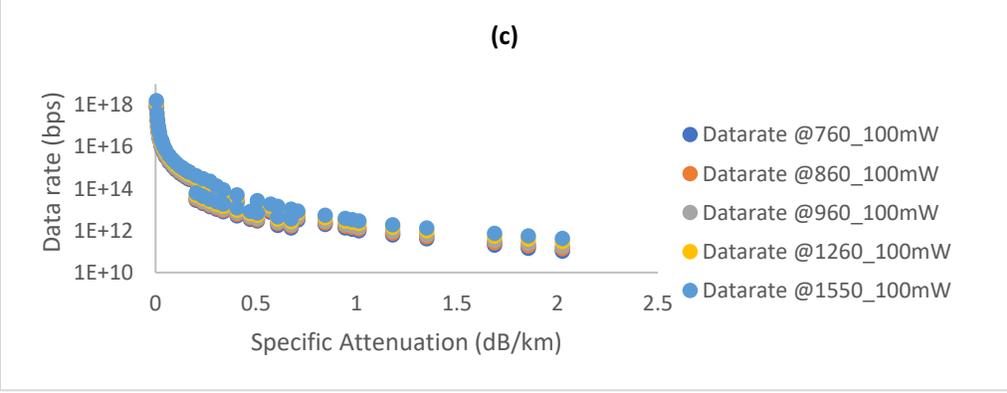

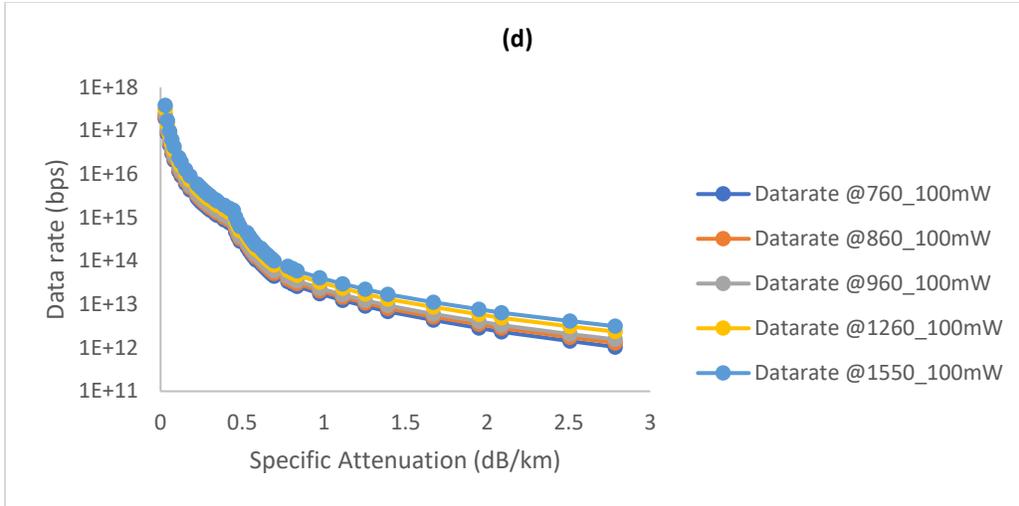

Figure 7: Data rate against specific attenuation at various wavelengths and 100 mW over **(a)** Polokwane (b) Kimberley (c) Bloemfontein (d) George

3.1.3 Power received against Link range

As presented in Figures 8(a-d), the link distances increase, as the optical received power reduces and this is in conformity with inverse square law. The reduction in optical received power may be as a result of misalignment between the transmitter and receiver due to atmospheric parameters like fog, dust etc. For an example, the maximum value of the power received at the wavelength of 1550 nm across the study locations Polokwane, Kimberley, Bloemfontein and George for an optical link range of 10 km is, 0.000555, 0.000681, 0.000680 and 0.000636 dBm respectively while at 760 nm at the same optical link range the maximum power received is, 0.000366, 0.000623, 0.00594 and 0.000524 dBm. This implies that FSO signal deploys at higher wavelength may tend to have better signal reception this is because optical power received increases with higher wavelengths.

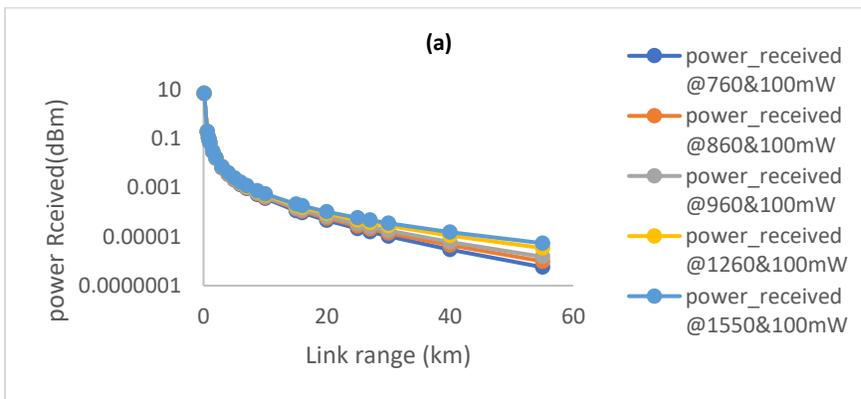

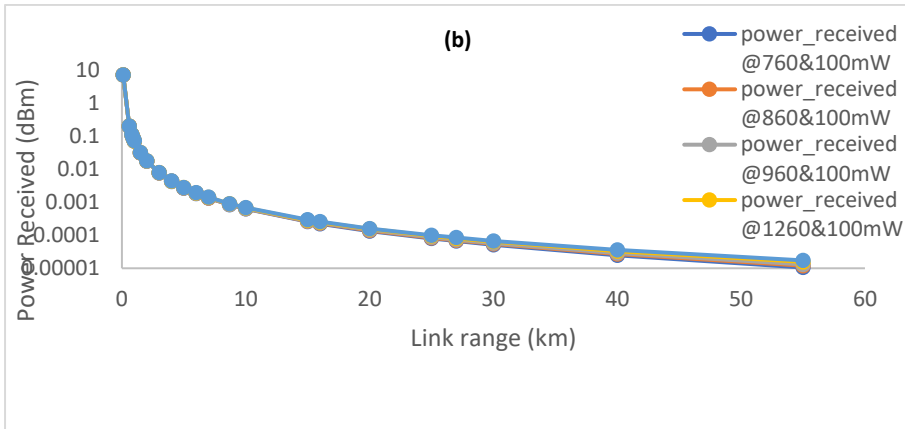

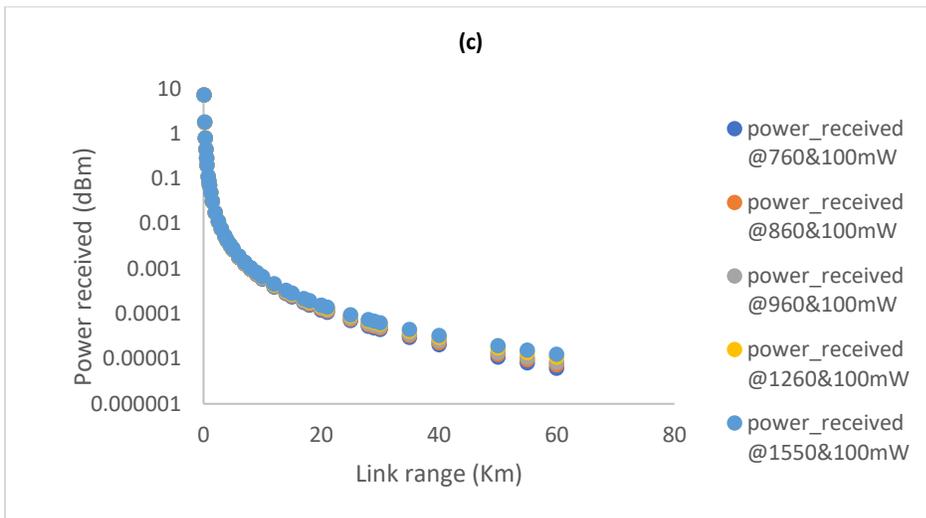

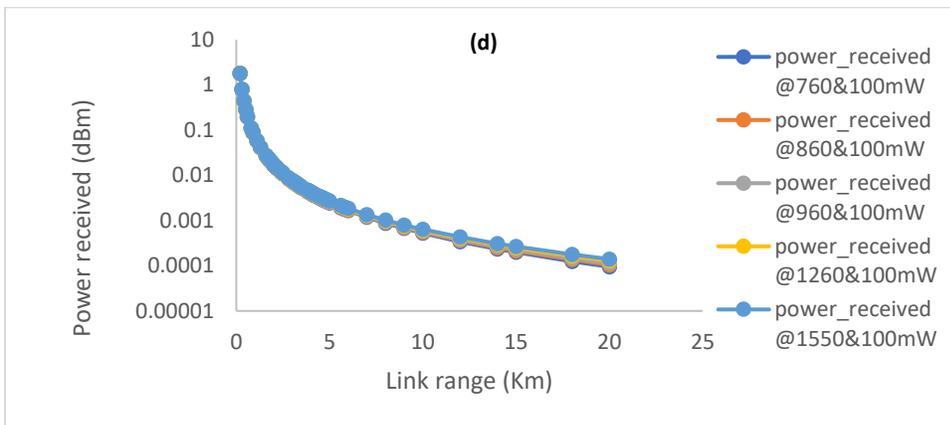

Figure 8 Power received against Link range at various wavelength and 100 mW over (a) Polokwane (b) Kimberley (c) Bloemfontein (d) George.

### 3.1.4 Evaluation of data rate and specific attenuation

As presented in Figures 9(a-d), the quality and dependability of the link are also described by the Bit Error Rate (BER), another distinctive communication component. It is also seen from the figures that increased attenuation will result in higher BER, which is inversely correlated with SNR. System data dependability would decline and BER could rise high if the signal has more noise power. The received signal quality will be low if the bit error rate is high. BER increases as attenuation increases. The FSO signal transmitted at 100 mw for 1 dB/km of attenuation, the BER values over the study locations: Polokwane, Kimberley, Bloemfontein, and George are, $6.743 \times 10^{-13}$, $6.730 \times 10^{-13}$, $6.750 \times 10^{-13}$, $and$ $6.755 \times 10^{-13}$ respectively.

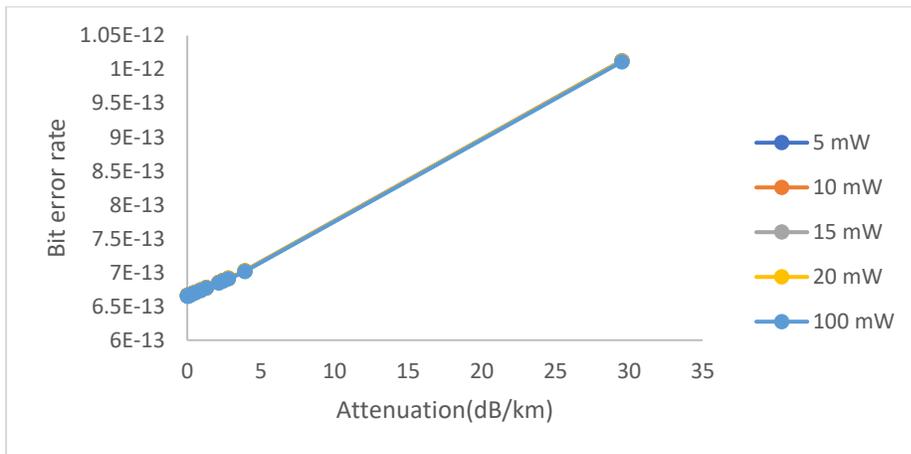

(a)

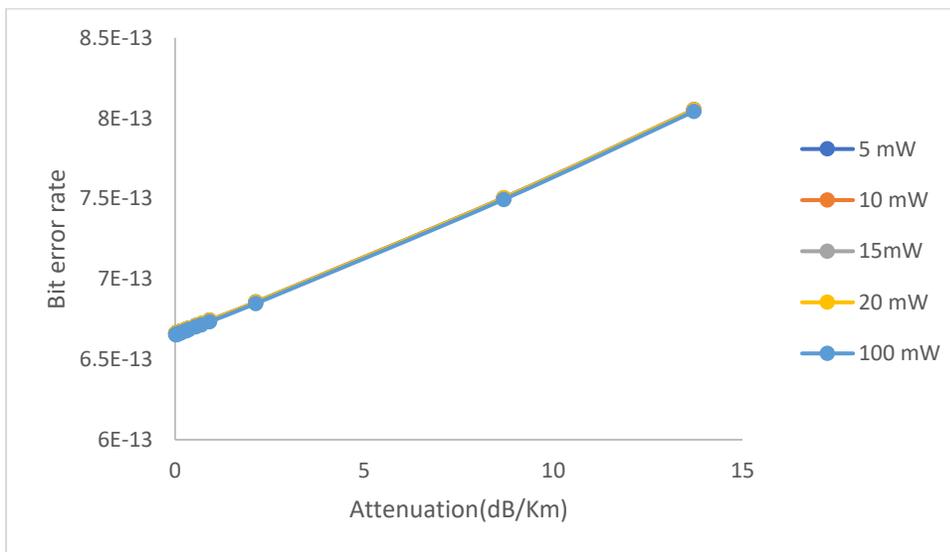

(b)

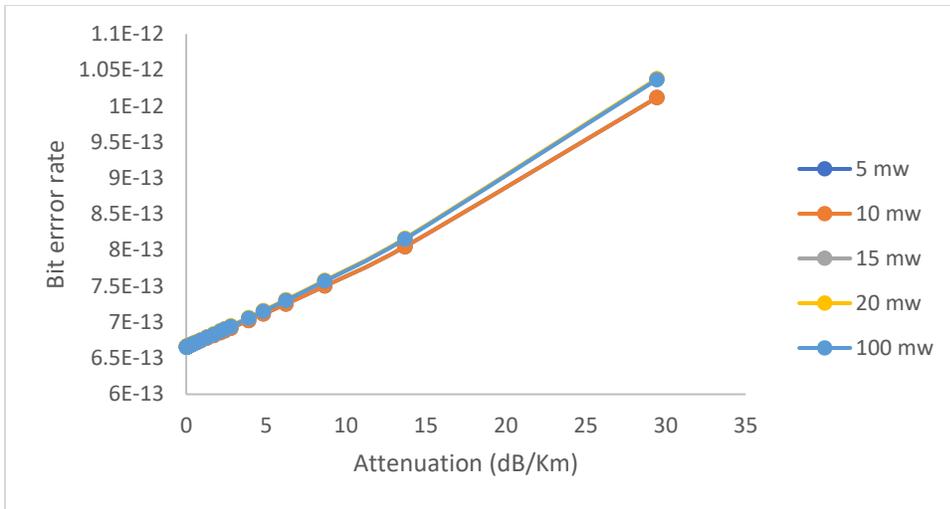

(c)

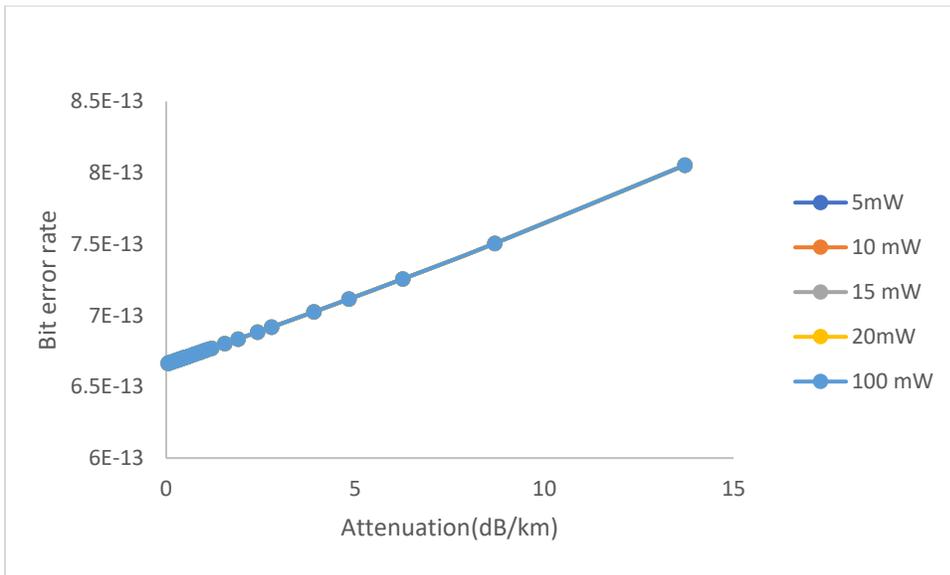

(d)

Figure 9 Bit error rate against Attenuation at various transmitted powers and 1550 nm over (a) Polokwane (b) Kimberley (c) Bloemfontein (d) George.

3.1.5 The variation of channel capacity and link distance

The figures 10(a-d), denote the influence of channel capacity on the link range at various wavelengths (1550, 1260, 960, 860, and 760 nm) over the study locations. It is observed that the channel capacity promptly hits a limit after beginning to rise steadily with link range. This pattern results from the Signal-to-Noise Ratio (SNR) being more stable with increasing link range. Capability may be impacted by the structure's increased SNR variations within a short range. The capacity achieves a stable level with increasing link range, exhibiting negligible variations. Also, the figures show channel stability after around 5 km in George while it is 10 km, in Polokwane, Bloemfontein and Kimberley. Although capacities at various wavelengths (1550 nm, 1260, 960, 860, and 760 nm, etc.) vary moderately, these figures display a negligible average variation. Higher wavelengths (such as 1550 nm) tend to work better for far-reaching FSO communication since they are less affected by meteorological factors like fog and rain. Because of its slightly greater SNR and reduced sensitivity to scattering under normal air circumstances, its capability at 1550 nm is slightly larger than the others. At a link distance of 1 km, the channel capacity values for a 1550 nm wavelength over the study locations, Polokwane, Kimberley, Bloemfontein, and George is 4.705Gbps.

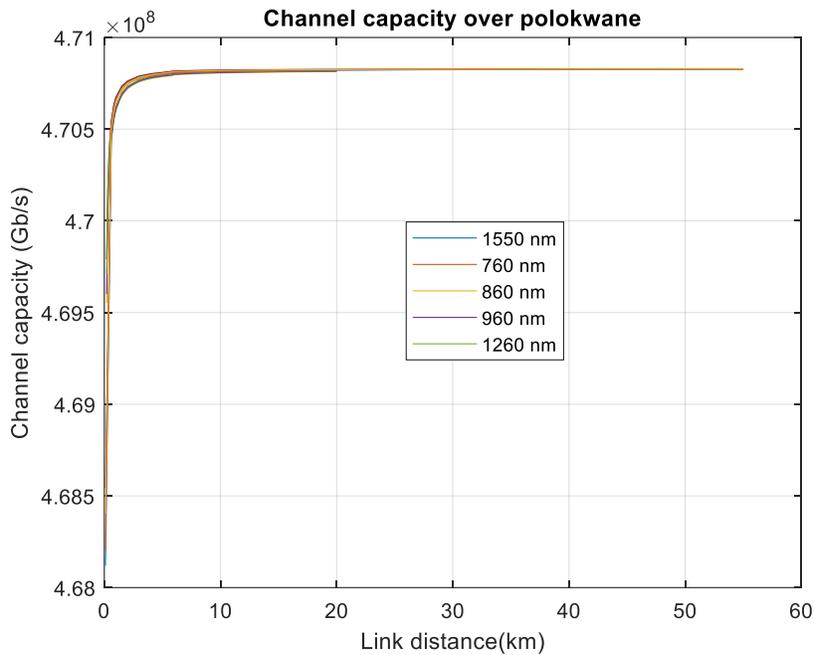

(a)

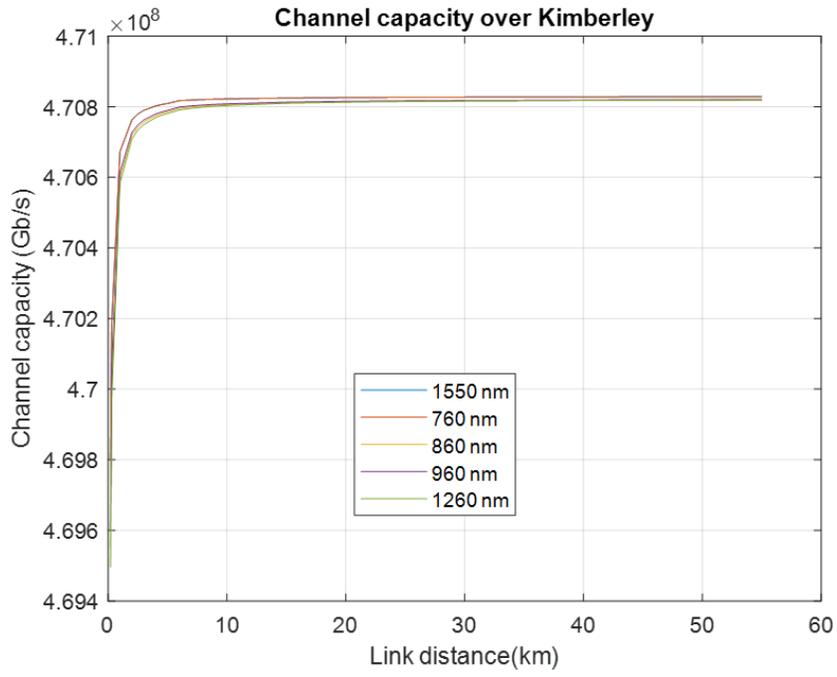

(b)

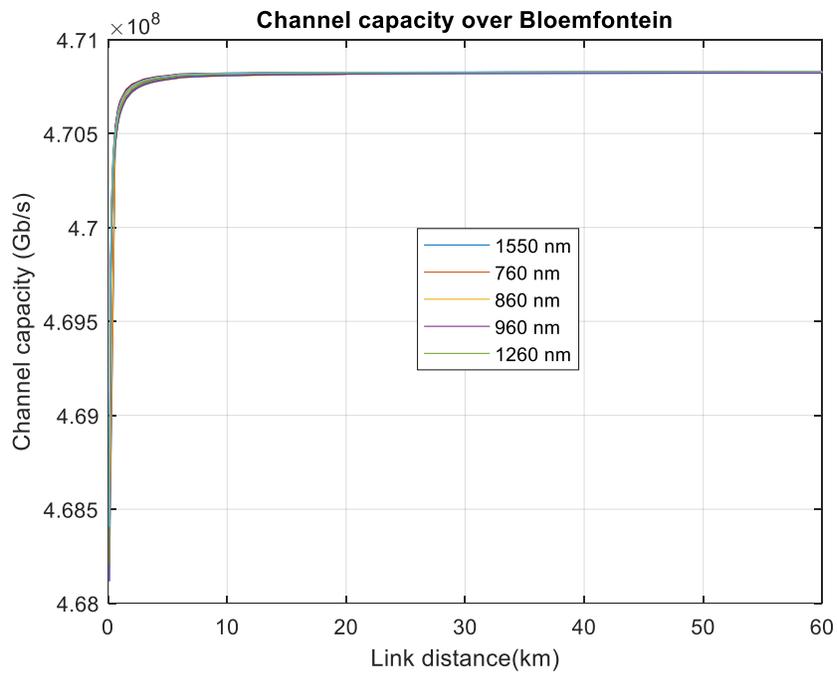

c)

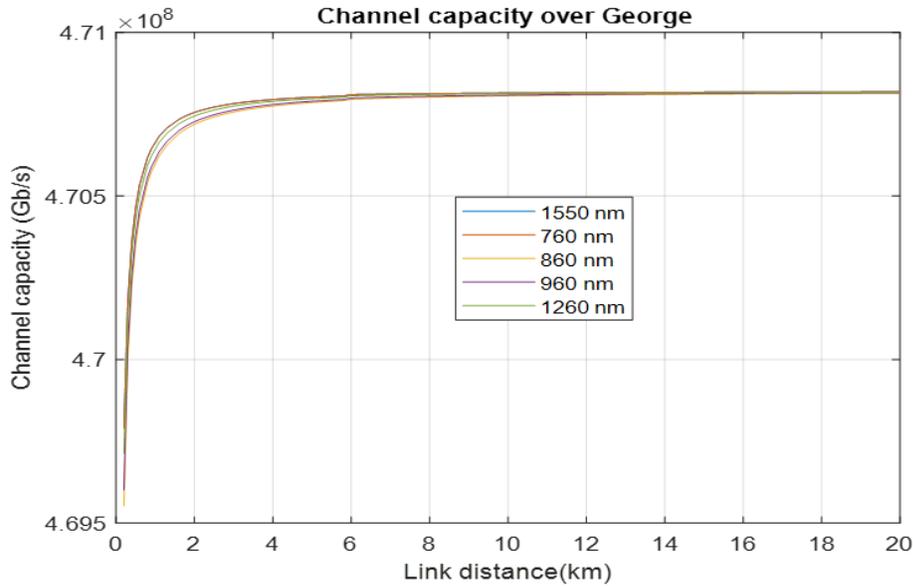

(d)

Figure10: Channel capacity against link distance at various wavelength over (a) Polokwane (b) Kimberley (c) Bloemfontein (d) George.

3.1.6 The variation of power penalty and link distance

The power penalty against link distance at a chosen BER of $10^{-9}$ is shown in figures11(a-d) under different fog conditions. According to the aforementioned investigation, the BER efficacy for the infiltration of noisy signals by dense fog into the predominant signal decreases with increasing connection distances in all the study locations. Therefore, additional power must be given to the local oscillator in order to properly differentiate the bits from the reception signal; this additional power makes up for the power penalty at that specific BER. It is also seen from all the figures that the dense fog conditions require high power compensation than other conditions across all the locations of study.

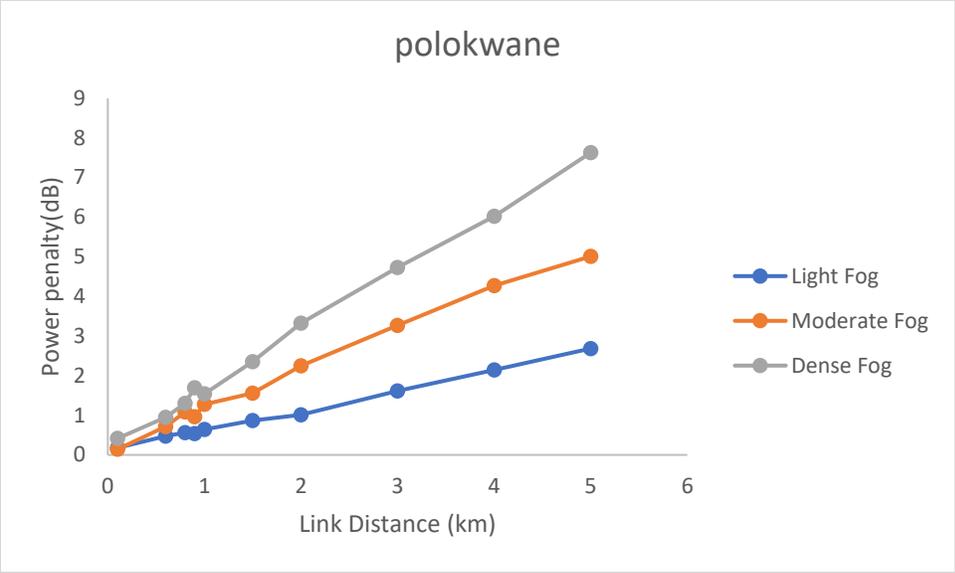

(a)

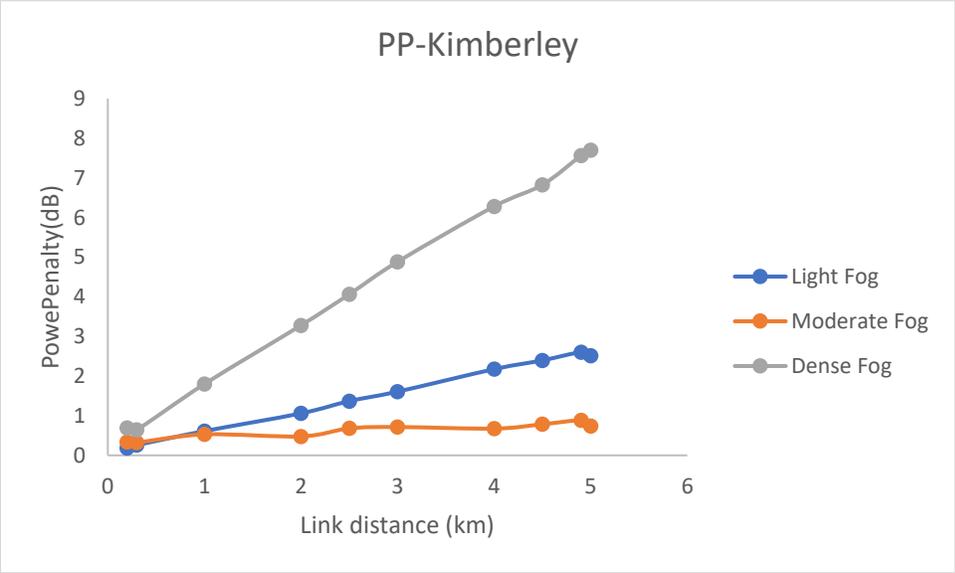

(b)

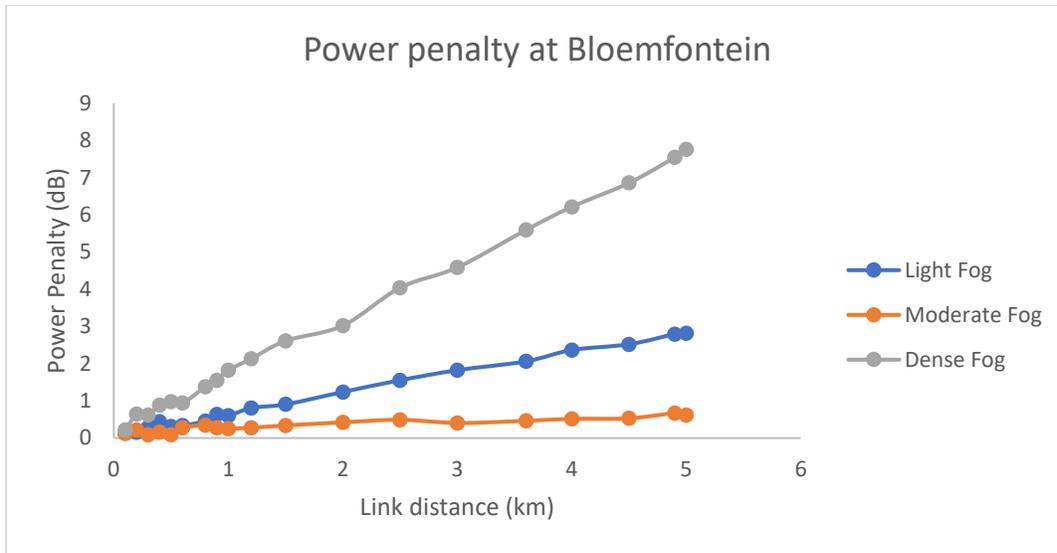

(c)

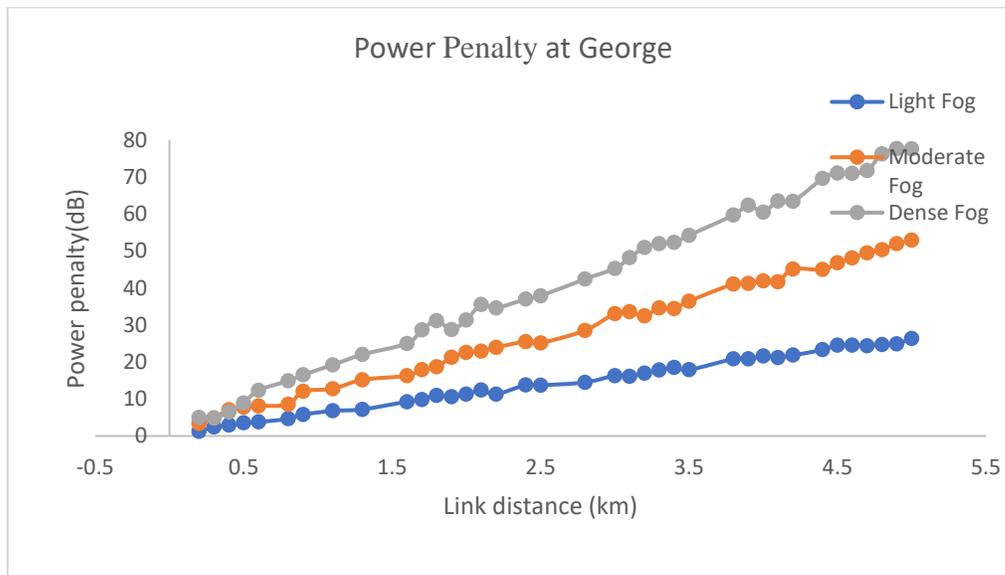

(d)

Figur11: Power penalty against link distance at various fog conditions over (a) Polokwane (b) Kimberley (c) Bloemfontein (d) George.

3.1.6 Comparison between the measured and the predicted QoS using various Ensemble learning Model

Figures 12 (a-d) show the comparison between the actual and forecast QoS over the study locations with different ensemble models. It is seen the figures that Multilayer neutral network could not account for the relationships that exist between the dependent variables and independent variables in predicting the QoS, hence, an erratic curve is displaced SR gives the best prediction of the measured QoS across all locations considered for this study.

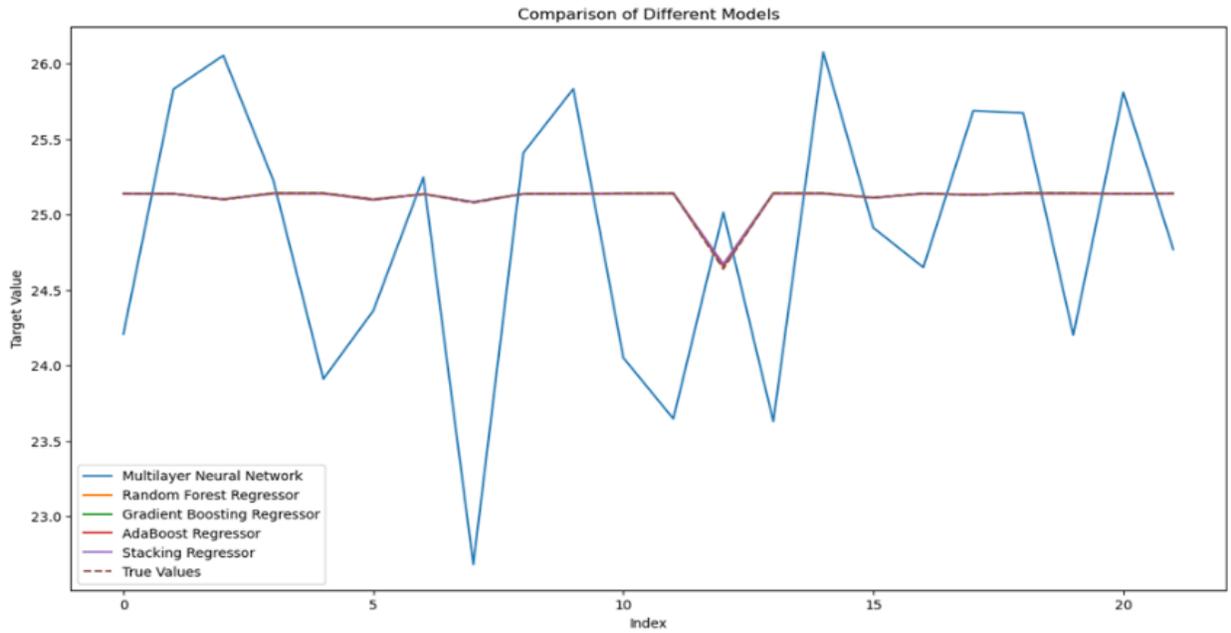

(a)

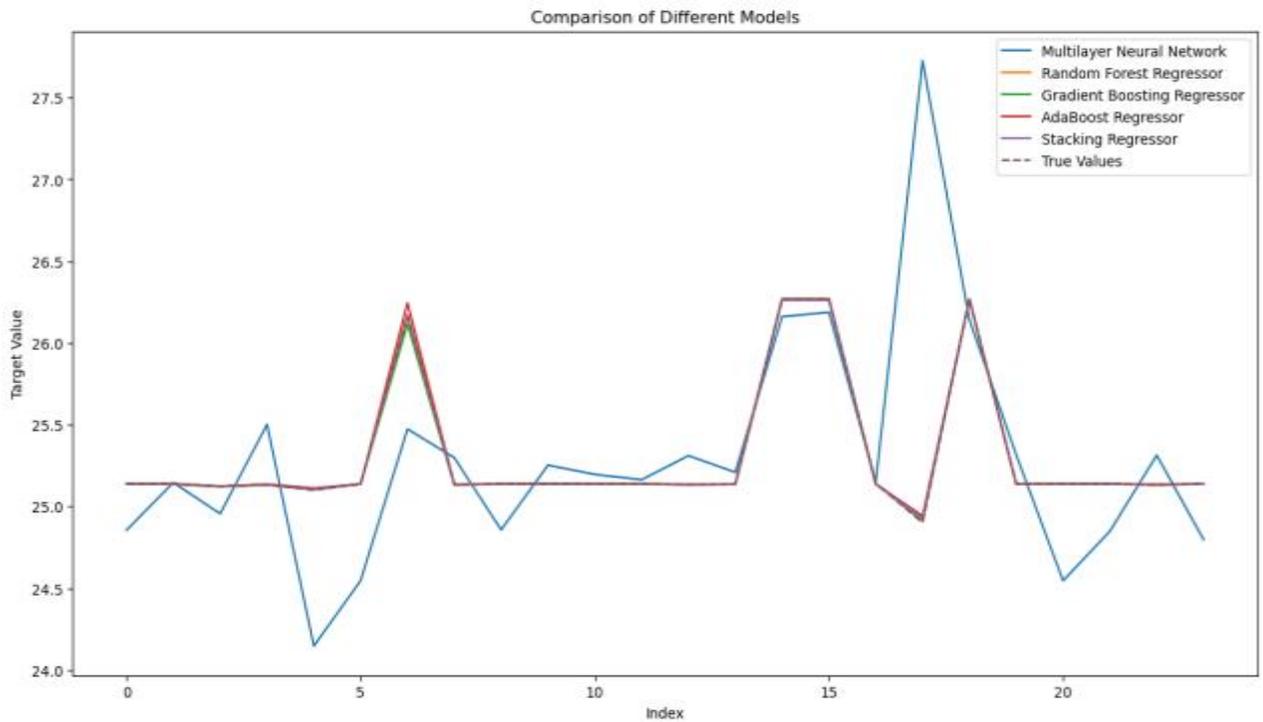

(b)

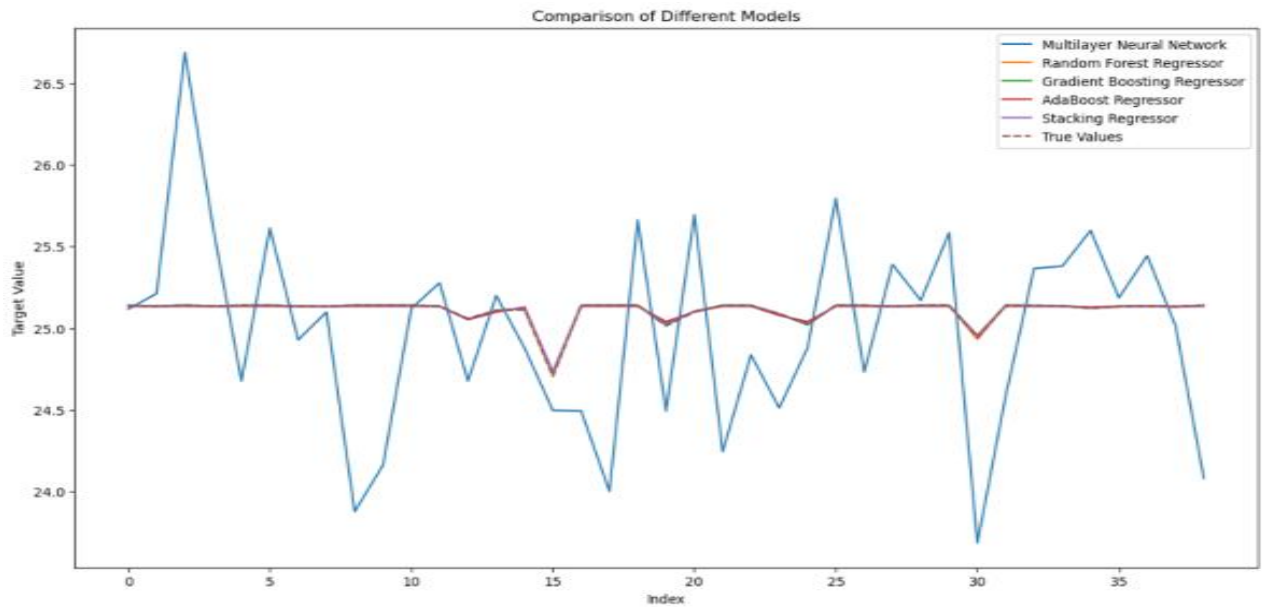

(c)

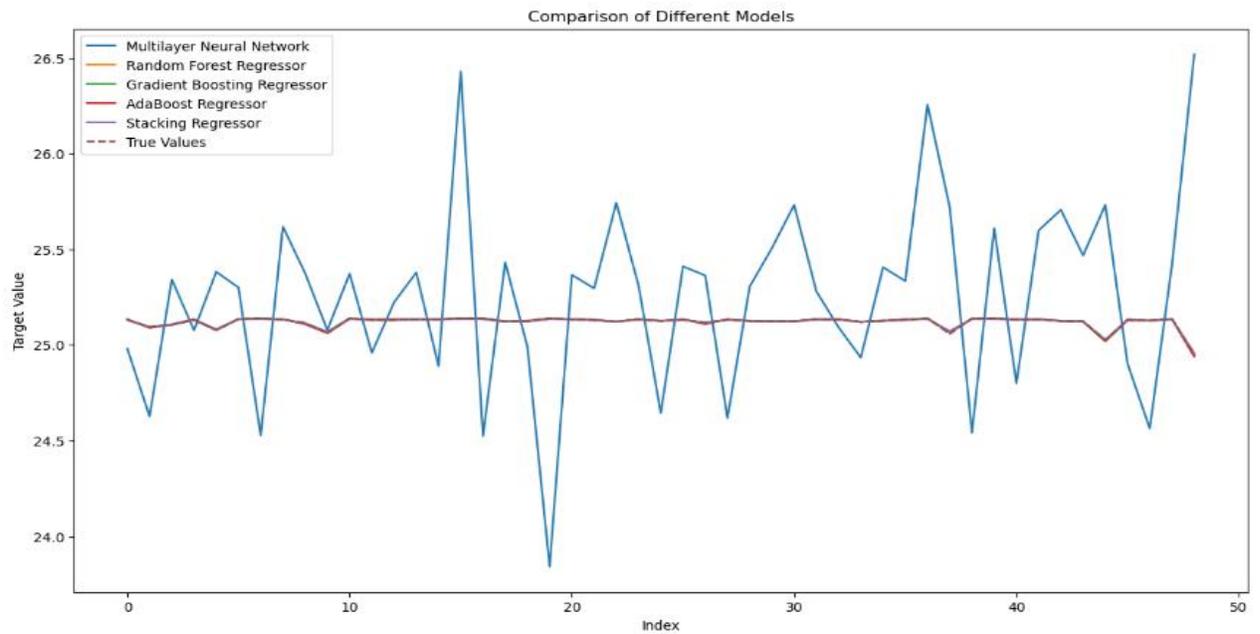

(d)

Figure12: Comparison of actual and forecast of QoS using different models over the study locations:(a) Polokwane (b) Kimberley (c) Bloemfontein (d) George.

3.5 Performance metric of various models over the studied locations.

Tables 3-6 show a comparison of the prediction performance of the five different models across the all the locations. According to table 3(a), the analysis of the fitting effect of the five models, the estimation performance reflected by MSE, MAE, MAPE, RMSE and $R^2$, in Polokwane is as follows: MLNN model (0.89602, 0.78531, 0.03127, 0.94658 and -81.6661); RF model (0.000048, 0.00212, 0.00008552, 0.00695, and 0.9955); GBR model (0.0000178, 0.00162, 0.00006533, 0.00421 and 0.9983); ADBR model (0.00002601, 0.00378, 0.0000151, 0.00051004; and 0.9976); SR model (0.0000529, 0.00335, 0.0000134, 0.00727; and 0.9951), GBR model has the minimum values of error and about 99.85% of R-squared value, this implies that the GBR model performs in predicting the QoS in Polokwane. For example, Table 3 shows that 99.85% variance of QoS could be accounted for by independent variables with the GBR model. Similarly, Table 4 presents the evaluation of performance metric of five different models in Kimberley as depicted by MSE, MAE, MAPE, RMSE and $R^2$, is as follows: MLNN model (0.4493, 0.3623, 0.0144, 0.0.6703 and -1.5360); RF model (0.00004181, 0.00219, 0.000087349, 0.006466, and 0.9998); GBR model (0.000136, 0.003291, 0.000127, 0.01167 and 0.9992); ADBR model (0.00031, 0.00727, 0.000285, 0.01744; and 0.9984); SR model (0.00004256, 0.003879, 0.000153, 0.006523; and 0.9996), SR model outperforms other models considered in this location because it has the minimum value error and also has R-squared value of 99.98%

Table 3: Performance metric of various models over Polokwane.

| Models | MSE | MAE | MAPE | RMSE | R^2 |
| --- | --- | --- | --- | --- | --- |
| MLNN | 0.89602 | 0.78531 | 0.031268 | 0.94658 | -81.6668 |
| RFR | 0.0000484 | 0.002119 | 0.0000855 | 0.00695 | 0.9960 |
| GBR | 0.0000178 | 0.00162 | 0.00000653 | 0.004218 | 0.9980 |
| ADBR | 0.0000260 | 0.0037784 | 0.000151 | 0.0051004 | 0.99759 |
| STR | 0.0000529 | 0.003346 | 0.0001343 | 0.00727 | 0.995117 |

Table 4: Performance metric of various models over Kimberley.

| Models | MSE | MAE | MAPE | RMSE | R^2 |
| --- | --- | --- | --- | --- | --- |
| MLNN | 0.44929 | 0.3632 | 0.01439 | 0.67039 | -1.53608 |
| RFR | 0.0000418 | 0.00219 | 8.73E-05 | 0.006466 | 0.99553 |
| GBR | 0.000136 | 0.00329 | 0.000127 | 0.01167 | 0.99923 |
| ADBR | 0.00031 | 0.00727 | 0.000285 | 0.01744 | 0.99838 |

| | STR | 0.0000426 | 0.003879 | 0.000153 | 0.006523 | 0.999759 |
|---|---|---|---|---|---|---|

Table 5: Performance metric of various models over Bloemfontein

| Models | MSE | MAE | MAPE | RMSE | R^2 |
|---|---|---|---|---|---|
| MLNN | 0.361 | 0.4622 | 0.01841 | 0.60083 | -52.86704 |
| RFR | 0.00000737 | 0.0015 | 0.0000587 | 0.00271 | 0.99950 |
| GBR | 0.0000258 | 0.00214 | 0.0000860 | 0.0051 | 0.99670 |
| ADBR | 0.0000783 | 0.0065 | 0.00026 | 0.00885 | 0.98784 |
| STR | 0.0000358 | 0.0036 | 0.000145 | 0.00598 | 0.99406 |

Table 6: Performance metric of various models over George.

| Models | MSE | MAE | MAPE | RMSE | R^2 |
|---|---|---|---|---|---|
| MLNN | 0.27429 | 0.40876 | 0.01627 | 0.52373 | -250.4 |
| RFR | 0.00000398 | 0.00103 | 0.0000410 | 0.001994 | 0.99600 |
| GBR | 0.00000804 | 0.00134 | 0.0000534 | 0.002834 | 0.99300 |
| ADBR | 0.00002030 | 0.0034 | 0.00014 | 0.00451 | 0.98136 |
| STR | 0.0000103 | 0.00219 | 0.0000871 | 0.0032 | 0.99100 |

Table 4 shows that 99.98% variance of QoS could be accounted for by independent variables with the SR model. Also, Table 5 illustrates the investigation of the fitting outcome of the five models, the estimation performance reflected by MSE, MAE, MAPE, RMSE and $R^2$, in Bloemfontein is as follows: MLNN model (0.3610, 0.4622, 0.01841, 0.60083 and -58.86704); RF model (0.00000737, 0.0015, 0.00005872, 0.00274, and 0.9987); GBR model (0.00002577, 0.00214, 0.00008595, 0.0051 and 0.99572); ADBR model (0.00007833, 0.0065, 0.000263, 0.00885; and 0.9870); SR model (0.0000358, 0.0036, 0.000145, 0.00598; and 0.99406), RF model outperforms other models considered in this location because it has the minimum value error and also has R-squared value of 99.87%. Table 5 shows that 99.87% variance of QoS could be accounted for by independent variables with the RF model. Also Table 6 presents the breakdown of the fitting result of the five models, the estimation performance reflected by MSE, MAE, MAPE, RMSE and $R^2$, in George is as follows : MLNN model (0.2743, 0.4088, 0.00163, 0.5237 and -250.4); RF model (0.00004, 0.00103, 0.00004100, 0.001994, and 0.9964); GBR model (0.00000804, 0.00134, 0.00005334, 0.00283 and 0.9923); ADBR model (0.00002033, 0.0034, 0.000014, 0.004451; and 0.9814); SR model (0.0000103, 0.00219, 0.00008714, 0.00320; and 0.9906), RF model outperforms other models considered in this location because it has the minimum value error and also has R-squared value of 99.64%. Table 6 shows that 99.64% variance of QoS could be accounted for by independent variables with the RF model. In the locations considered for this study, the MLNN performs poorly.

4. Conclusion and future work

To mitigate the transmission channel fading situation, this study has provided an ensemble technique that forecasts channel impairment owing to sub-tropical climate variables. The method selects the suitable propagating variables automatically. In this study, it was found that higher QoS might be generated to satisfy customers' Service Level Agreements (SLAs) by using Ensemble for signal modeling. The capacity to produce a single-valued result will culminate in satisfactory results. Concerning different transmitted wavelengths, it was discovered that the fog-induced attenuation of greater dB/km was observed at a lower optical link range across all the study locations, Polokwane, Kimberley, Bloemfontein, and George. However, the loss of signal caused by fog decreases with increasing link distance. The findings indicate that the diminution impact lessens with increasing visibility as the operating optical wavelength increases. Besides, the evaluation of the findings shows that as the propagated optical wavelengths increase, it leads to the improvement in data rate, and power received. The findings demonstrated that, incorporating an ensemble learning techniuqe for transmission modeling can effectively create a higher quality of service (QoS) to meet customers' service level agreements (SLAs). This will make it possible to produce a single worthwhile product, leading to an effective performance. In the end, the combined result will provide FSO systems engineers with the technologies required to achieve the required link margin for optimal service quality during periods of intense subtropical weather conditions

Acknowledgement

The Authors also like to thank South African Weather Services for providing weather data for the study and Tshwane University of Technology Pretoria, South Africa.